\documentclass[acmsmall]{acmart}

\usepackage{graphicx}
\usepackage[caption=false,font=footnotesize]{subfig}
\usepackage{multirow}

\usepackage{algorithm}
\usepackage{algpseudocode}

\usepackage{tcolorbox} 
\tcbuselibrary{skins} 

\newtcolorbox{mybox}{
  colback=gray!20, 
  colframe=gray!40, 
  boxrule=0.5pt,
  boxsep=2pt,
  arc=4pt,
  left=3pt, 
  right=3pt,
  top=3pt,
  bottom=3pt
}
\usepackage{enumitem}

\AtBeginDocument{%
  }

\setcopyright{cc}
\setcctype{by}
\acmDOI{10.1145/3715722}
\acmYear{2025}
\acmJournal{PACMSE}
\acmVolume{2}
\acmNumber{FSE}
\acmArticle{FSE005}
\acmMonth{7}
\received{2024-09-11}
\received[accepted]{2025-01-14}

\begin{document}

\title{On-Demand Scenario Generation for Testing Automated Driving Systems}

\author{Songyang Yan}
\orcid{0000-0003-0628-8642}
\affiliation{%
  \institution{Xi'an Jiaotong University}
  \city{Xi'an}
  \country{China}
}
\email{tayyin@stu.xjtu.edu.cn}

\author[1]{Xiaodong Zhang}
\authornote{Corresponding authors}
\orcid{0000-0002-8380-1019}
\affiliation{%
  \institution{Xidian University}
  \city{Xi'an}
  \country{China}
}
\email{zhangxiaodong@xidian.edu.cn}

\author{Kunkun Hao}
\orcid{0000-0003-0420-1347}
\affiliation{%
  \institution{Synkrotron}
  \city{Xi'an}
  \country{China}
}
\email{haokunkun@synkrotron.ai}

\author{Haojie Xin}
\orcid{0000-0001-5223-0113}
\affiliation{%
  \institution{Xi'an Jiaotong University}
  \city{Xi'an}
  \country{China}
}
\email{pinkman@stu.xjtu.edu.cn}

\author{Yonggang Luo}
\orcid{0009-0000-3973-7606}
\affiliation{%
  \institution{Chongqing Changan Automobile}
  \city{Chongqing}
  \country{China}
}
\email{luoyg3@changan.com.cn}

\author{Jucheng Yang}
\orcid{0000-0002-5046-2663}
\affiliation{%
  \institution{Chongqing Changan Automobile}
  \city{Chongqing}
  \country{China}
}
\email{yangjc3@changan.com.cn}

\author{Ming Fan}
\orcid{0000-0002-9327-0987}
\affiliation{%
  \institution{Xi'an Jiaotong University}
  \city{Xi'an}
  \country{China}
}
\email{mingfan@mail.xjtu.edu.cn}

\author{Chao Yang}
\orcid{0000-0002-3311-291X}
\affiliation{%
  \institution{Xidian University}
  \city{Xi'an}
  \country{China}
}
\email{chaoyang@xidian.edu.cn}

\author{Jun Sun}
\orcid{0000-0002-3545-1392}
\affiliation{%
  \institution{Singapore Management University}
  \city{Singapore}
  \country{Singapore}
}
\email{junsun@smu.edu.sg}

\author[1]{Zijiang Yang}
\authornotemark[1]
\orcid{0009-0002-5437-0253}
\affiliation{%
  \institution{University of Science and Technology of China}
  \city{Hefei}
  \country{China}
}
\affiliation{%
  \institution{Synkrotron}
  \city{Xi'an}
  \country{China}
}
\email{yang@synkrotron.ai}

\renewcommand{\shortauthors}{S. Yan, X. Zhang, K. Hao, H. Xin, Y. Luo, J. Yang, M. Fan, C. Yang, J. Sun, and Z. Yang}

\begin{abstract}
The safety and reliability of Automated Driving Systems (ADS) are paramount, necessitating rigorous testing methodologies to uncover potential failures before deployment. Traditional testing approaches often prioritize either natural scenario sampling or safety-critical scenario generation, resulting in overly simplistic or unrealistic hazardous tests. In practice, the demand for natural scenarios (e.g., when evaluating the ADS's reliability in real-world conditions), critical scenarios (e.g., when evaluating safety in critical situations), or somewhere in between (e.g., when testing the ADS in regions with less civilized drivers) varies depending on the testing objectives. To address this issue, we propose the On-demand Scenario Generation (OSG) Framework, which generates diverse scenarios with varying risk levels. Achieving the goal of OSG is challenging due to the complexity of quantifying the criticalness and naturalness stemming from intricate vehicle-environment interactions, as well as the need to maintain scenario diversity across various risk levels. OSG learns from real-world traffic datasets and employs a Risk Intensity Regulator to quantitatively control the risk level. It also leverages an improved heuristic search method to ensure scenario diversity. We evaluate OSG on the Carla simulators using various ADSs. We verify OSG's ability to generate scenarios with different risk levels and demonstrate its necessity by comparing accident types across risk levels. With the help of OSG, we are now able to systematically and objectively compare the performance of different ADSs based on different risk levels. 
\end{abstract}



\maketitle

\section{Introduction}

\begin{figure}[h]
\centerline{\includegraphics[width=0.9\textwidth]{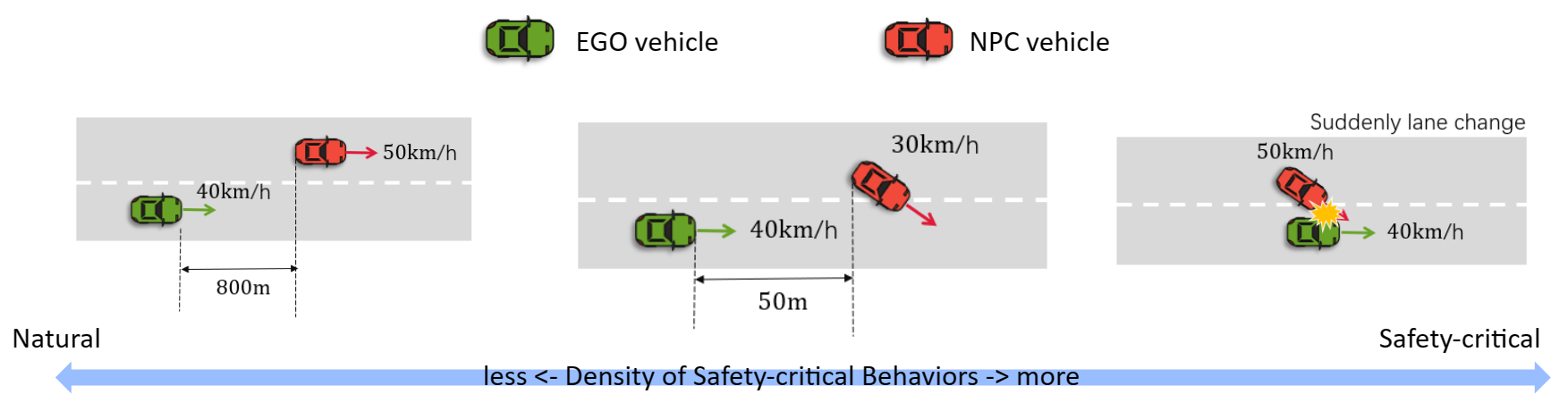}}
\caption{Comparison of scenarios at different risk levels. The green vehicle is the ADS under test (i.e., EGO), and the red vehicle is the background vehicle (i.e., NPC). \textbf{Left scenario:} The EGO and NPC are slowly moving in different lanes, with a far distance between them. \textbf{Middle scenario:} a slower NPC attempts to merge into the lane occupied by the EGO, forcing the EGO to yield. \textbf{Right scenario:} An adjacent NPC suddenly changes lanes into the EGO's lane, resulting in a collision.}
\label{nat_adv_compare}
\end{figure}

Ensuring safety is crucial for Automated Driving Systems (ADS). Recent incidents, such as a Tesla in Autopilot mode failing to detect a crossing truck, highlight this need ~\cite{Thadani_Lerman_Piper_Siddiqui_Uraizee_2023}. The National Highway Traffic Safety Administration (NHTSA) has documented 736 crashes involving Tesla vehicles equipped with Autopilot, resulting in 17 fatalities since 2019 ~\cite{Siddiqui_Merrill_2023}. Testing is vital for safety assurance. It typically includes real-vehicle and simulation tests, with the latter preferred for its cost-effectiveness and zero risk.

Testing ADS requires a diverse set of driving scenarios with varying levels of risk. According to a survey of autonomous driving test engineers ~\cite{lou2022testing}, the common practice involves using data from real-world driving scenarios and accident reports to comprehensively test ADS. 
However, the testing process faces significant challenges due to the lack of a standardized method for generating scenarios with different risk levels.

Firstly, the testing requirements for ADS vary depending on the specific objectives of the tests. In some cases, testers need natural scenarios that closely resemble real-world driving conditions to assess the system's performance in everyday situations. In other cases, critical scenarios that push the limits of the ADS are necessary to identify potential weaknesses and edge cases. Additionally, scenarios with intermediate levels of risk are crucial for evaluating the system's behavior in moderately challenging conditions. A generator capable of creating scenarios with various levels of risks and naturalness is essential to meet these diverse testing needs.
Figure ~\ref{nat_adv_compare} demonstrates that scenarios with different risk levels possess distinct values. The first scenario, a natural one that frequently occurs in real life, features an NPC (non-player character) vehicle that poses almost no threat to the EGO (ADS under test) vehicle, and maintaining safe driving is the baseline for the ADS. In the second scenario, which has a medium risk level, a slower NPC cuts in closely in front of the EGO, requiring the EGO to yield. The final scenario, characterized by a high-risk level, involves a high-speed cut-in by an NPC, making it nearly impossible for the EGO vehicle to avoid a collision. This scenario can be used to test the ADS's performance under extreme conditions, assess its capability limits, and determine whether it can minimize losses.

Secondly, without a standardized method for measuring the criticalness and naturalness of the scenarios, it is difficult to properly compare the safety of different ADSs, or different versions of a developing ADS. For example, comparing the number of crashes through random testing can be misleading since the randomly generated scenarios may have different levels of naturalness. 
To address this issue, a consistent and objective way of quantifying the criticalness and naturalness of scenarios is necessary.  This would enable testers to generate scenarios with desired characteristics and make fair comparisons during their development of ADS.

Recent studies have aimed at automating scenario generation, typically using three methods: data-driven~\cite{scanlon2021waymo, abdelfattah2021towards, chen2021geosim, savkin2021unsupervised}, knowledge-based~\cite{shiroshita2020behaviorally, wang2021commonroad,  Chang2024LLMScenarioLL, Woodlief2024S3CSS}, and adversarial generation approaches~\cite{zhu2021adversarial, ding2021multimodal, ghodsi2021generating, arief2021deep, koren2018adaptive, cai2020summit, Wang2024DanceOT}. Data-driven methods often produce natural but repetitive scenarios due to the data's long-tail effect. Knowledge-based approaches, relying on human-designed rules, may lack efficiency and struggle with diversity. In contrast, adversarial generation methods, known for their efficiency and flexibility, are increasingly popular. They treat scenario generation as a search problem. However, many of these methods focus either on scenario naturalness or criticalness, failing to offer flexibility in the controlling the two aspects.

To generate scenarios that meet the test requirements,
we proposed the On-demand Scenario Generation (OSG) Framework, capable of generating scenarios with different risk levels.  Grading the risk level of scenarios allows testers to objectively test and compare different ADSs. Testers can select scenarios with different risk levels based on the development stage of the autonomous driving system and specific needs. 

Generating diverse driving scenarios with varying risk levels presents several challenges. 
Firstly, quantifying the criticalness and naturalness of scenarios is complex due to the intricate interactions between vehicles and the environment.
Secondly, even if there is a way of measuring criticalness and naturalness, we still need to design an efficient searching procedure to avoid generating duplicate scenarios which wastes testing resources and reduce the effectiveness of the testing results.

To address the first challenge, we designed a Naturalness Estimator and a Risk Indicator to quantify the naturalness of a scenario and its criticalness. We employ generative models to learn from traffic datasets. These data are gathered from the real world and encompass a broad spectrum of natural traffic behavior patterns. The trained model assesses the naturalness of newly generated scenarios. We extract the temporal relative states of the ADS concerning other vehicles within the scenario, calculating the overall risk level of the scenario based on their kinematic parameters. These two modules are regulated through a Risk Intensity Regulator (RIR), guiding the generator to produce scenarios of varying risk levels.

To address the second challenge, we use a speciation-based optimization method to search for scenarios. This heuristic search method simulates the process of biological evolution into different species, categorizing the generated scenarios into various groups. This approach enables the discovery of multiple local optima in the parameter space, i.e.,  it can generate a diverse set of scenarios.

To evaluate the effectiveness of the proposed OSG Framework, we conducted experiments on the Carla simulator ~\cite{Dosovitskiy17} and highway-env ~\cite{highway-env} using various ADSs, including Baidu Apollo ~\cite{apollo}, end-to-end agents (Interfuser ~\cite{shao2023safety} and TF++ ~\cite{jaeger2023hidden}), and rule-based agents ( Carla Behavior Agent ~\cite{Dosovitskiy17} and highway agent ~\cite{highway-env}). 
We evaluated the scenario set generated by OSG using several widely used indicators, verifying that OSG can generate scenarios with different risk levels as required. 
We compared OSG with the state-of-the-art scenario generator BehAVExplor ~\cite{cheng2023behavexplor}. The evaluation results show that OSG discovers more accidents (an increase of 92.97\% on average) than the baseline.
We further demonstrate the necessity of OSG by comparing the types of accidents encompassed in scenarios at different risk levels.
We also designed a comprehensive ADS scoring system based on OSG to rate the safety of different ADSs and different versions of Apollo. The results show that our method can objectively reflect the progress of an ADS.

In summary, the main contributions of our work are:

1. The development of the OSG framework, which can generate diverse traffic scenarios with varying risk levels for testing ADSs.

2. The evaluation of OSG's effectiveness in identifying defects within diverse ADS architectures, including rule-based, end-to-end, and hybrid systems.

3. The proposal of a comprehensive scoring method based on OSG-generated scenarios to quantitatively assess the safety performance of ADS.

\section{Illustrative Example}

\begin{figure}[t]
\centerline{\includegraphics[width=0.7\textwidth]{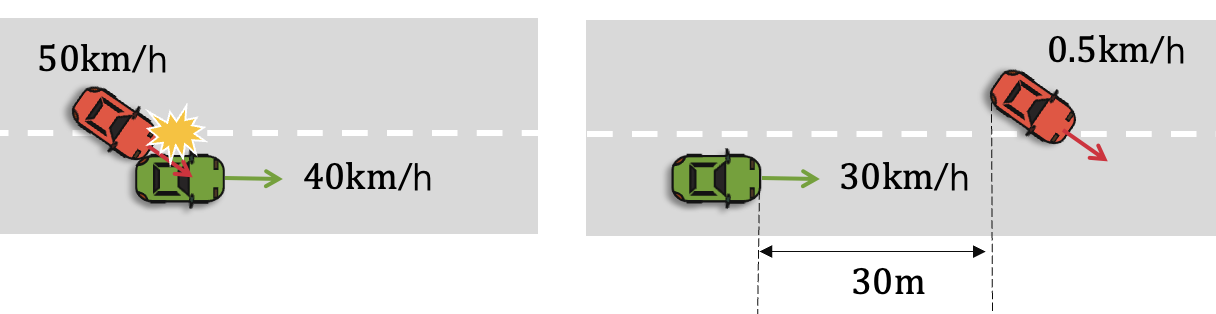}}
\caption{Highlighted scenarios in the example test suite.}
\label{example_case}
\end{figure}

\begin{figure}[t]
\centerline{\includegraphics[width=1\textwidth]{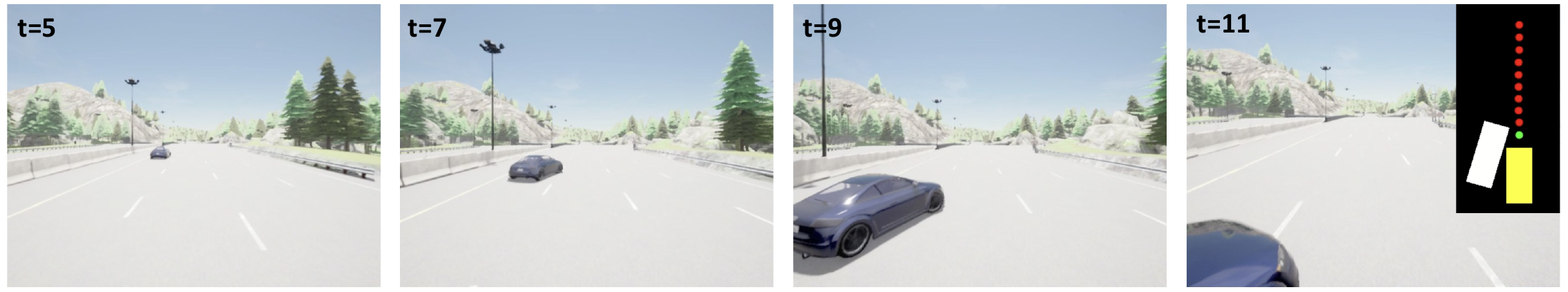}}
\caption{In the Carla simulator, the Interfuser agent did not detect a slow-moving NPC encroaching into the lane, resulting in a collision. The four screenshots, arranged from left to right, capture the Carla simulator's state at 5, 7, 9, and 11 seconds of simulation time. The last image highlights the wire-frame representation of the vehicles at the point of collision.
}
\label{example_interfuser_crash}
\end{figure}


Let us assume that a tester aims to conduct a comprehensive assessment of multiple ADSs in a highway environment.
The test scenario can be summarized as follows: on a two-lane highway, the EGO vehicle travels on the right lane, with an NPC vehicle on the left lane, which may change lanes to the right at any moment. 


Many methods generate test suites by data sampling ~\cite{fang2020augmented, manivasagam2020lidarsim, scanlon2021waymo} or by adversarial search ~\cite{tivhkk, ding2021multimodal, wang2021advsim, sun2022lawbreaker} based on a series of risk metrics. However, the risk levels of these scenarios are either too low or too high, not meeting the comprehensive testing needs of ADS. OSG can generate scenarios of different risk levels in batches.

Utilizing the OSG framework, the tester can obtain a test suite of different risk levels for this scenario and use OSG's comprehensive scoring system for a quantitative assessment of ADSs.

In the test suite generated by OSG in this example, two cases are highlighted as shown in Figure ~\ref{example_case}. The left one is the highest risk level scenario: the NPC changes lanes abruptly when in the blind spot of the EGO, directly hitting the EGO. In this scenario, it is impossible for the EGO to react in time. For testers, such testing results are not useful, i.e., while it is critical, it is not something natural to be handled by ADS.

Another is a medium-risk level scenario: the NPC changes lanes very slowly, merely moving a small part of the vehicle body out of the lane line, then waits in place. The following EGO vehicle incorrectly estimated the extent of the NPC's lane intrusion, did not slightly shift to the right to avoid it, and thus collided with the NPC vehicle. This might indicate precision errors in EGO's sensors or planning system. This scenario exposes risks not covered by high-risk level scenarios. Such scenarios genuinely exist when testing Interfuser ~\cite{shao2023safety}, an end-to-end ADS, as shown in Figure ~\ref{example_interfuser_crash}, which demonstrates the scenario in the Carla simulator where Interfuser collides with an almost stationary NPC.

Based on the scoring system provided by OSG, we evaluated the safety performance of the rule-based agent Highway Agent ~\cite{highway-env} and the industrial-level ADS Baidu Apollo ~\cite{apollo}. The results show that Apollo scored 85.06, while Highway Agent scored 61.65, indicating that Apollo has better safety performance.
However, if we only use natural scenarios to evaluate the ADS, the scores are 84.35 for Apollo and 86.21 for Highway Agent, suggesting that Highway Agent has higher safety performance than Apollo. Such an evaluation however is potentially misleading. Although rule-based agents perform more robustly in non-hazardous environments, they cannot handle complex traffic situations.

In summary, OSG can uncover more potential defects in ADSs through scenarios with different risk levels. It can also provide reasonable safety performance scores for ADSs based on these diverse scenarios.

\section{Proposed Method}

\begin{figure}[t]
\centerline{\includegraphics[width=1\textwidth]{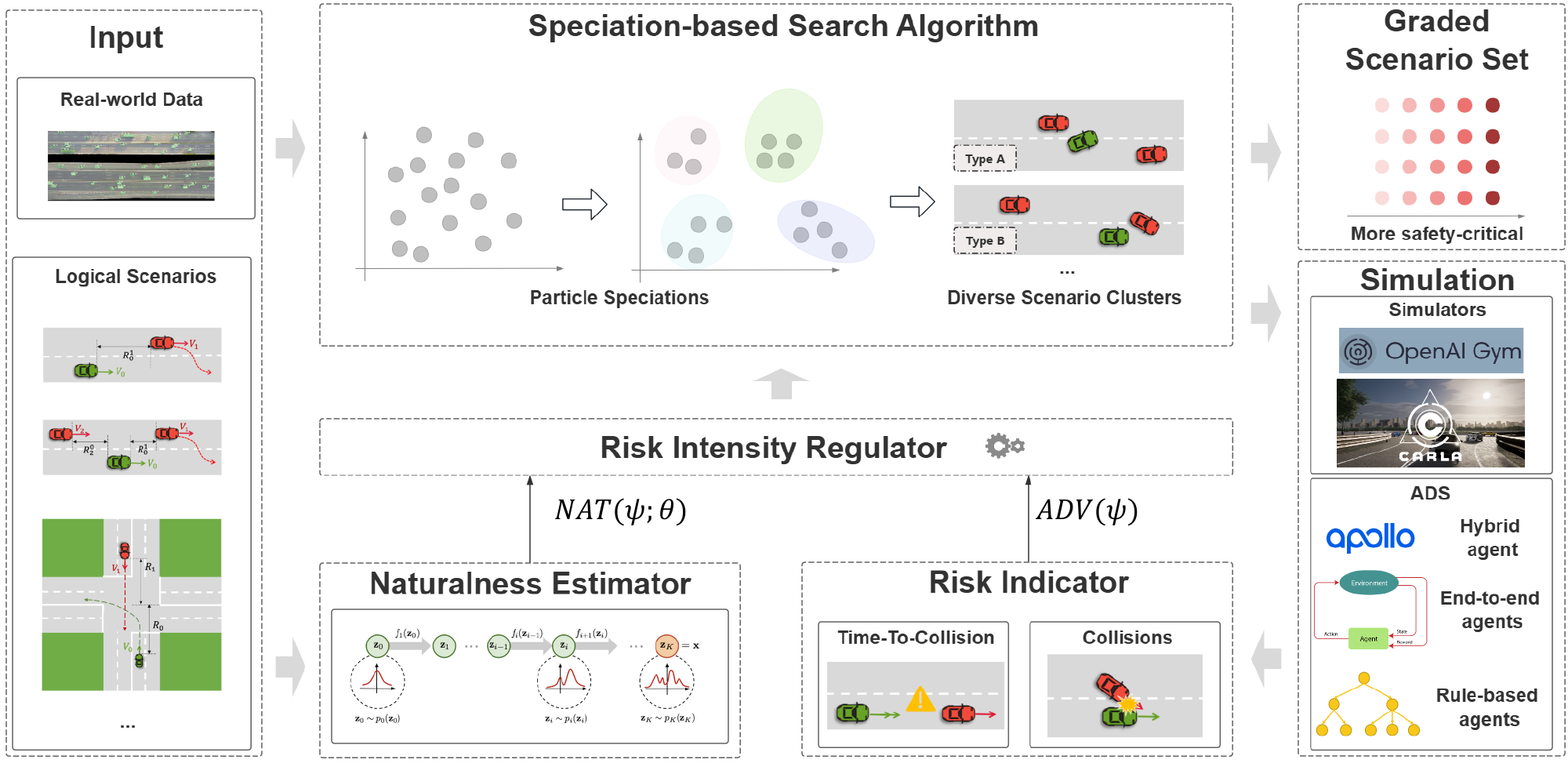}}
\caption{The overall framework of our method for generating traffic scenarios.}
\label{overall_framework}
\end{figure}

Figure \ref{overall_framework} illustrates the architecture of the OSG framework. The input to OSG consists of two parts: realistic traffic datasets and scenario specifications. From these inputs, we train a Naturalness Estimator which quantitatively evaluates the naturalness of newly generated scenarios. To produce scenarios with varying risk levels according to the test requirements, OSG also defines a Risk Indicator. This indicator measures the risk level of scenarios using two quantitative metrics: Time-To-Collision ~\cite{minderhoud2001extended} and the number of collisions. The outputs from both the Naturalness Estimator and the Risk Indicator form the Risk Intensity Regulator, which serves as a reward function during scenario generation.

OSG frames the scenario generation problem as an optimization problem. It employs a searching algorithm to find suitable combinations of scenario parameters within the parameter space. We design a specific optimization algorithm to efficiently locate diverse local optima, preventing the creation of many homogeneous scenarios.

The generated scenarios are then simulated on various combinations of simulators and ADSs. A unified data collector gathers simulation data, calculates various metrics, and provides feedback to the scenario search module. Ultimately, OSG outputs a set of scenarios with different risk levels, tailored to the testing needs.

\subsection{Scenario Representation}

\begin{figure}[t]
\centerline{\includegraphics[width=1\textwidth]{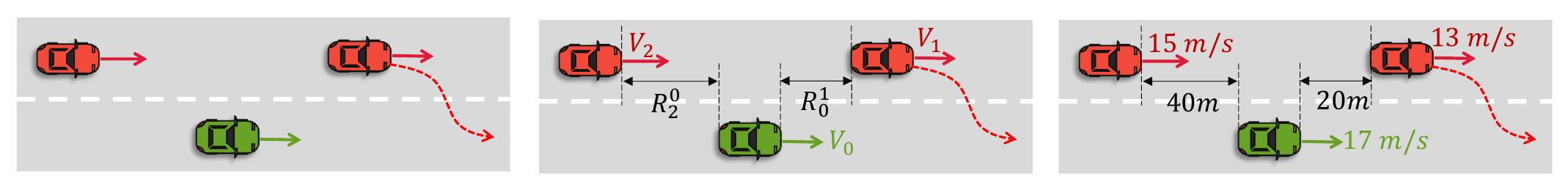}}
\caption{Three stages of a scenario. From left to right, they are the functional scenario, logical scenario, and concrete scenario. }
\label{scen_desc}
\end{figure}

Traffic scenarios can be organized into a three-tiered hierarchy: functional scenarios (FS), logical scenarios (LS), and concrete scenarios (CS). An FS provides a high-level description of a scenario. For instance, the FS visualized in Figure ~\ref{scen_desc} can be described as follows: An EGO vehicle is hindered by a cut-in vehicle with another vehicle in the adjacent left lane.

The LS adds further details to the FS by identifying a set of critical variables, $\Psi$, and defining their value ranges, providing quantitative measures required for generating the scenario.
\begin{equation}
LS = \{ (\psi_i, [a_i, b_i]) \mid \psi_i \in \Psi, a_i, b_i \in \mathbb{R}, a_i \leq b_i \}.
\end{equation}
In this process, each $\psi_i$ represents a specific scenario parameter, and $[a_i, b_i]$ denotes the allowable value range for $\psi_i$.

The CS is derived from the LS by assigning specific values to each parameter identified in the LS, thereby creating a set of precise parameter-value pairs that fully concretize the scenario:
\begin{equation}
CS = \{ (\psi_i, x_i) \mid \psi_i \in \Psi, x_i \in [a_i, b_i] \subset \mathbb{R} \}.
\end{equation}

\subsection{Traffic Prior Model}

OSG adopts a systematic parameterization of traffic scenarios, which is essential to our analysis. To effectively evaluate the similarity between the generated scenarios and real traffic conditions, we need to model real traffic datasets.

We construct a model of highway driving using the Next Generation Simulation (NGSIM) \cite{usdot2016ngsim} dataset, which offers detailed vehicle trajectories from various US locations. This dataset enables the study of common highway behaviors like lane changes and car-following. For urban intersection scenarios, we used the Interaction \cite{zhan2019interaction} and In-D \cite{inDdataset} datasets, which include diverse driving situations at intersections, roundabouts, and merging lanes.



    
    


Let $T$ be the set of vehicle trajectory data points, where each $t \in T$ is a tuple $t = (c, s, l, f_{id}, r_{id}, \tau)$, where $c$ represents the coordinates, $s$ the speed, $l$ the lane ID, $f_{id}$ the front vehicle ID, $r_{id}$ the rear vehicle ID, and $\tau$ the time step.

Given a logical scenario $LS$ defined by a set of intervals for each parameter, we define a predicate function $\phi: T \times LS \to \{0, 1\}$. This function evaluates to 1 if a data point $t$ satisfies the conditions of $LS$ within a specified time window $\Delta \tau$ around a critical event, and 0 otherwise.

We can then formalize the extraction process as a mapping $E: T \times LS \to 2^T$, which selects a subset of trajectory data points from $T$ that satisfy $LS$:
\begin{equation}
E(T, LS) = \{ t \in T \mid \phi(t, LS) = 1 \}.
\end{equation}

We define an MAF (Masked Autoregressive Flow) model \cite{papamakarios2017masked} $M$ as a sequence of autoregressive transformations that learn a mapping from the data distribution to a Gaussian distribution.
MAF leverages a series of reversible autoregressive transformations to map the data distribution to a Gaussian distribution. The advantage of MAF lies in its ability to capture complex dependencies within the data while maintaining a simple and efficient training process. MAF is particularly well-suited for probability density estimation tasks, as it can explicitly compute the likelihood probabilities of data points.
The model $M$ is parameterized by a set of weights and biases, collectively denoted as $\theta$. The training process of $M$ involves adjusting $\theta$ to maximize the likelihood of the data points in $E(T, LS)$.

Let $f_M: \mathbb{R}^n \to \mathbb{R}$ be the density estimation function of the MAF model $M$, which takes a vector of scenario parameters $\psi$ as input and outputs the likelihood of $\psi$ under the learned distribution:
$ f_M(\psi; \theta) = \text{likelihood of } \psi \text{ under the model } M. $

For each Logical Scenario, we train a unique MAF model to estimate the naturalness of the corresponding Concrete Scenarios.

\subsection{Risk Intensity Regulator}

The scenario generation problem is formalized as a search-based optimization problem that aims to maximize an objective function $G(\psi; \omega, \theta)$. This function, calculated by the Risk Intensity Regulator, controls the risk level of the scenarios. The objective function is defined as follows:
\begin{equation}
\label{eq-refined-opt}
\max_{\psi \in \Psi} G(\psi; \omega, \theta) = \max_{\psi \in \Psi}  \left( \widetilde{ADV}(\psi)^{\omega^2} + \widetilde{NAT}(\psi;\theta)^{(1-\omega)^2} \right)^{e^{\omega(1-\omega)}} ,
\end{equation}
where $\Psi$ represents the parameter space of a given logical scenario, and $\psi \in \Psi$ denotes a concrete scenario within this space.

The objective function $G(\psi; \omega, \theta)$ comprises two parts: a base function and an exponential modifier. In the base function, the balance weight $\omega \in [0, 1]$ modulates the relative importance of the adversarial and naturalness terms. As $\omega$ approaches $0$, the optimization prioritizes the naturalness of scenarios, while as $\omega$ approaches $1$, it prioritizes the criticalness presented by the scenarios. The exponential term $e^{w(1-w)}$ serves as a scaling factor that amplifies the effect of the weighting parameter $w$.

The criticalness of a scenario $\psi$ is quantified by $ADV: \Psi \rightarrow \mathbb{R}$, which is given by:
\begin{equation}
\label{eq-refined-adv}
ADV(\psi) =  C(\psi) \cdot P_{\text{col}} - minTTC(\psi) ,
\end{equation}
where $minTTC(\psi)$ computes the minimum value among the Time-to-Collision (TTC) between the ADS and other vehicles at every instant during the scenario $\psi$. $C(\psi)$ represents the number of collisions that the ADS being tested has had with other vehicles in scenario $\psi$. The term $P_{\text{col}} \in \mathbb{R}^+$ is a predefined collision reward coefficient.

The naturalness term, $NAT(\psi;\theta): \Psi \times \Theta \rightarrow \mathbb{R}$, is a function parameterized by $\boldsymbol{\theta} \in \Theta$, which represents the parameters of a density estimation model. It outputs the log-likelihood of scenario $\psi$ under the estimated distribution of real-world data.

In Equation ~\ref{eq-refined-opt}, $\widetilde{ADV}$ and $\widetilde{NAT}$ are normalized version of $ADV(\psi)$ and $NAT(\psi;\theta)$. Their range is $[0,1]$.

\subsection{Scenario Generation}

\begin{algorithm}[t]
\caption{Speciation-based Particle Swarm Optimization for Scenario Generation}
\label{algo-spso}
\begin{algorithmic}[1]
\Require Logical scenario $LS$, risk parameter $\omega$, population size $N$, maximum iterations $M$, speciation threshold $\tau$
\Ensure A set of scenarios grouped by different species $\mathcal{S}$
\State Initialize population $\mathcal{P} = \{\mathbf{P_1}, \mathbf{P_2}, \ldots, \mathbf{P_N}\}$
\For{$t = 1$ to $M$}
    \For{each particle $\mathbf{P_i} \in \mathcal{P}$}
        \State Evaluate $G(\mathbf{P_i})$ using $G(\psi; \omega)$
        \State Update $\mathbf{P_i^{pbest}}$ if $G(\mathbf{P_i}) > G(\mathbf{P_i^{pbest}})$
    \EndFor
    \State Perform speciation on $\mathcal{P}$ based on $\tau$
    \For{each species $\mathcal{S}_j$}
        \State Update $\mathbf{S_j^{sbest}}$ as the best $\mathbf{P_i^{pbest}}$ in $\mathcal{S}_j$
    \EndFor
    \For{each particle $\mathbf{P_i} \in \mathcal{P}$}
        \State Update velocity $\mathbf{V_i}$ using $\mathbf{P_i^{pbest}}$ and $\mathbf{S_j^{sbest}}$
        \State Update $\mathbf{P_i} = \mathbf{P_i} + \mathbf{V_i}$
        \State Constrain $\mathbf{P_i}$ within $LS$ ranges
    \EndFor
\EndFor
\State \Return $\mathcal{S}$
\end{algorithmic}
\end{algorithm}

We approach the problem of scenario generation by transforming it to an optimization problem, where the Risk Intensity Regulator controls the objective. 
We adopt a speciation-based particle swarm optimization (SPSO) algorithm ~\cite{brits2002niching}. The choice of SPSO is motivated by its unique incorporation of the concept of species, which significantly reduces the risk of overfitting during the search process. Moreover, SPSO has a propensity to yield a greater variety of scenarios.
Generating diverse scenarios is of paramount importance for the rigorous testing of ADSs. A multitude of similar scenarios may repeatedly expose a particular system flaw.

We analogize a scenario to a particle within an optimization manifold. The manifold's dimensionality $D = |\Psi|.$
The optimization endeavor involves generating a subset of scenarios $\mathcal{S} \subset \mathbb{R}^D$.
We define the threshold $\tau_i$:
$$ \tau_i = \frac{b_i - a_i}{C^{1/D}}. $$
$b_i$ and $a_i$ are respectively the upper and lower bounds of the value of $\psi_i \in \Psi$. $C$ is a constant.

For speciation classification, we consider two particles $\mathbf{P_1}$ and $\mathbf{P_2}$ in $\mathbb{R}^D$ as members of different species if there exists at least one dimension $i$ in the set $\{1, \ldots, D\}$ for which the following disparity holds:
$$ \exists i \in \{1, \ldots, D\} \text{ such that } |\mathbf{P_1}(i) - \mathbf{P_2}(i)| > \tau_{i}. $$
Here, $\mathbf{P_1}(i)$ and $\mathbf{P_2}(i)$ represent the $i$-th components of the respective particles.

Algorithm ~\ref{algo-spso} presents the scenario generation process. It takes a logical scenario $LS$, risk parameter $\omega$, and several configurable parameters as inputs, and outputs a set of scenarios $\mathcal{S}$ that are divided into several groups.

The algorithm initializes a population $\mathcal{P}$ of particles within the search space defined by $LS$ (line 1). It then repeats the following steps until the maximum number of iterations $M$ is reached (lines 2-16). In each iteration, it evaluates the objective function $G(\psi; \omega, \theta)$ for each particle $\mathbf{P_i}$ and updates the personal best position $\mathbf{P_i^{pbest}}$ if the current position improves the objective function value (line 5). The algorithm performs speciation on the population $\mathcal{P}$ based on the threshold $\tau$ (line 7) and updates the species best position $\mathbf{S_j^{sbest}}$ for each species $\mathcal{S}_j$ (lines 8-10). It then updates each particle's velocity $\mathbf{V_i}$ and position $\mathbf{P_i}$ using $\mathbf{P_i^{pbest}}$ and $\mathbf{S_j^{sbest}}$ (lines 11-15) while ensuring that the particles remain within the feasible region defined by $LS$ (line 14). Finally, the algorithm returns a set of scenarios grouped by different species (lines 17).

This algorithm explores the scenario space effectively by maintaining multiple species during the optimization process. It prevents premature convergence to a single local optimum and encourages the generation of diverse scenarios.

\section{Experiments}

In this section, we evaluate the effectiveness of OSG in generating different risk-level scenarios. More specifically, we answer the following research questions:

\begin{itemize}[label={},leftmargin=*]
    \item \textbf{RQ1: }Is it necessary to test ADS in scenarios with varying levels of risk?
    \item \textbf{RQ2: }Can OSG effectively generate scenarios across a spectrum of risk levels?
    \item \textbf{RQ3: }How effective is OSG in identifying defects within diverse ADS architectures?
\end{itemize}

To answer the above research questions, we implemented the OSG framework both in the lightweight simulator highway-env 1.8.2 ~\cite{highway-env} based on OpenAI GYM ~\cite{towers_gymnasium_2023} and the high-fidelity Carla 0.9.14 simulator ~\cite{Dosovitskiy17}. In the selection of test subjects, we choose 5 ADSs, listed in Table ~\ref{agents-used}. Among them, the Highway Agent runs on highway-env, and the other agents run on Carla.

\begin{table}[t]
  \caption{Agetns used in experiments. Interfuser and TF++ do not have version numbers. The latest commit time of their GitHub repository is used instead.}
  \centering
  \label{agents-used}
\begin{tabular}{@{}lll@{}}
\toprule
Agent                                & Type               & Version    \\ \midrule
Apollo ~\cite{apollo}                & Multi-module-mixed & 9.0        \\
Interfuser ~\cite{shao2023safety}    & End-to-end         & 2024.1.20  \\
TF++ ~\cite{jaeger2023hidden}        & End-to-end         & 2023.12.14 \\
Highway Agent ~\cite{highway-env}    & Rule-based         & 1.8.2      \\
Behavior Agent ~\cite{Dosovitskiy17} & Rule-based         & 0.9.14     \\ \bottomrule
\end{tabular}
\end{table}

\begin{figure}[t]
    \centering
    \subfloat[FrontBrake]{\includegraphics[width=0.25\textwidth]{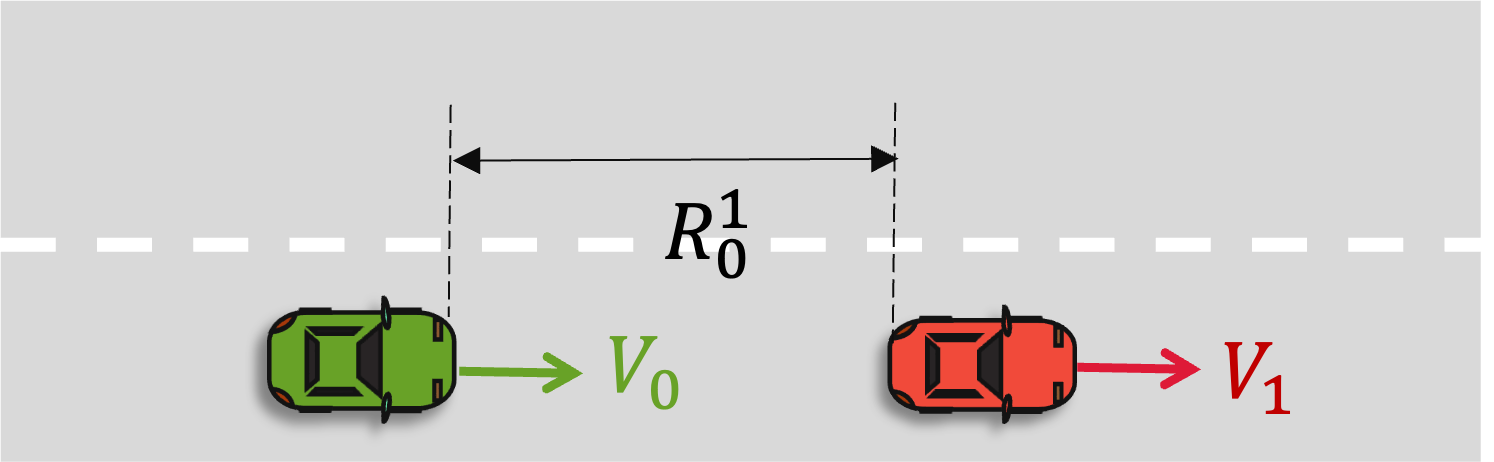} \label{scen-fb} }
    \hspace{0.03\textwidth}
    \subfloat[CutIn-1]{\includegraphics[width=0.25\textwidth]{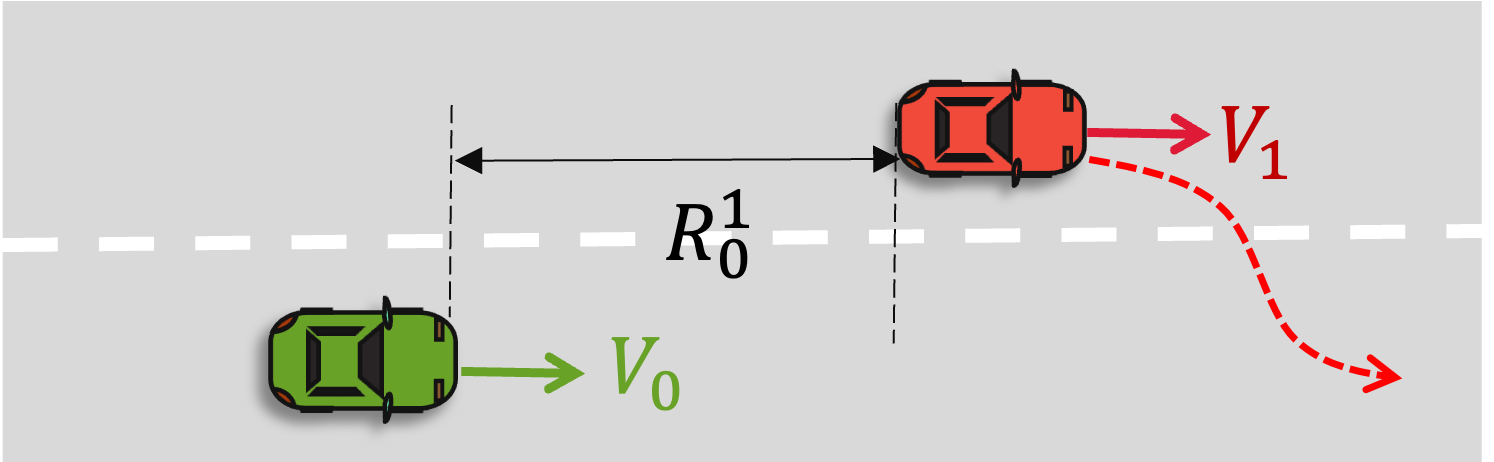}}
    \hspace{0.03\textwidth}
    \subfloat[CutIn-2]{\includegraphics[width=0.25\textwidth]{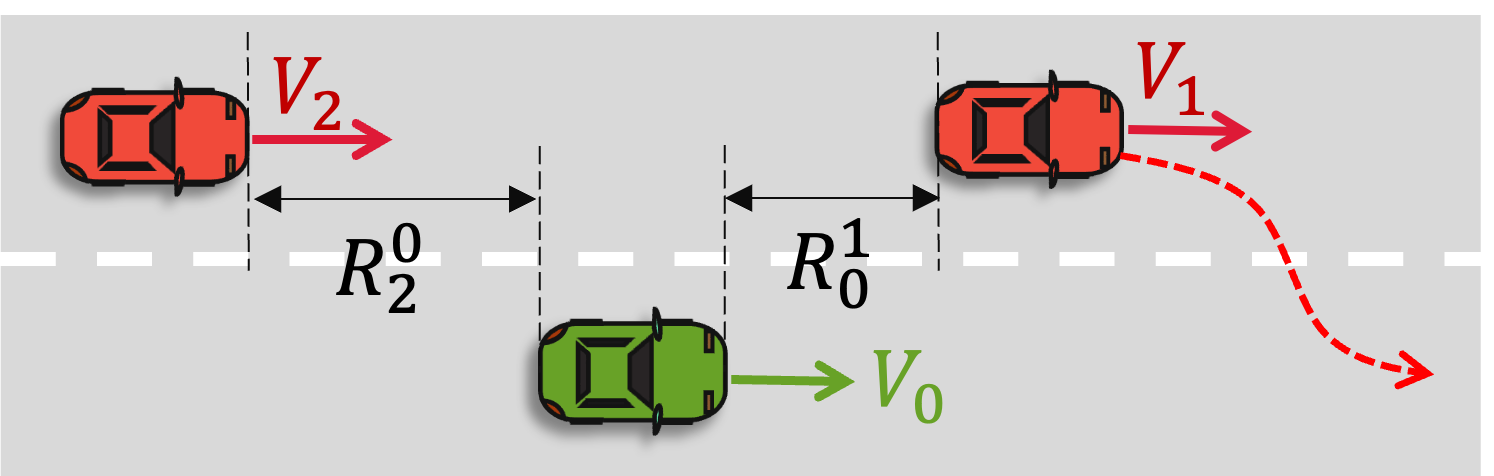}}
    \\
    \subfloat[OVTP]{\includegraphics[width=0.25\textwidth]{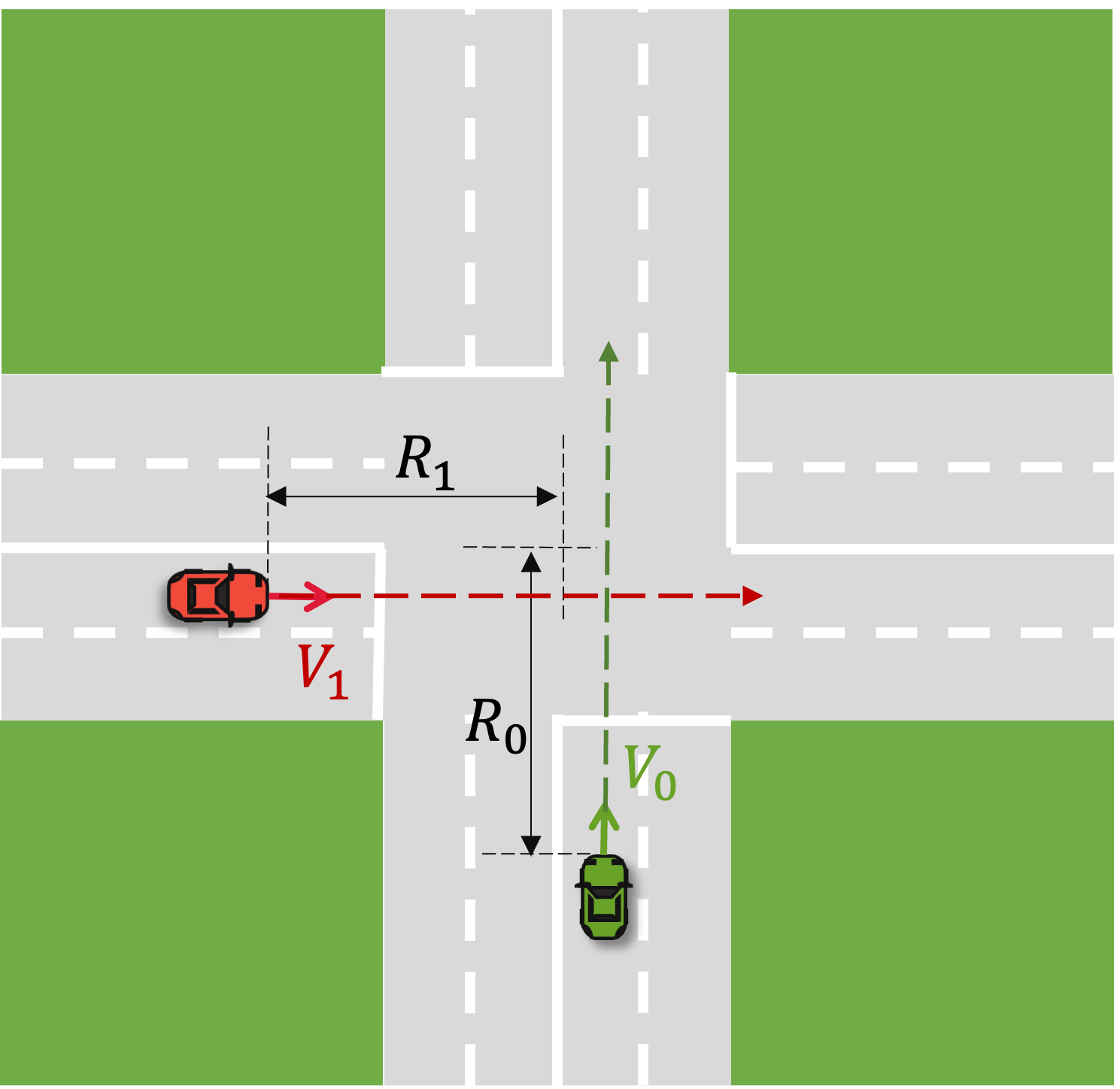}}
    \hspace{0.03\textwidth}
    \subfloat[NJLT]{\includegraphics[width=0.25\textwidth]{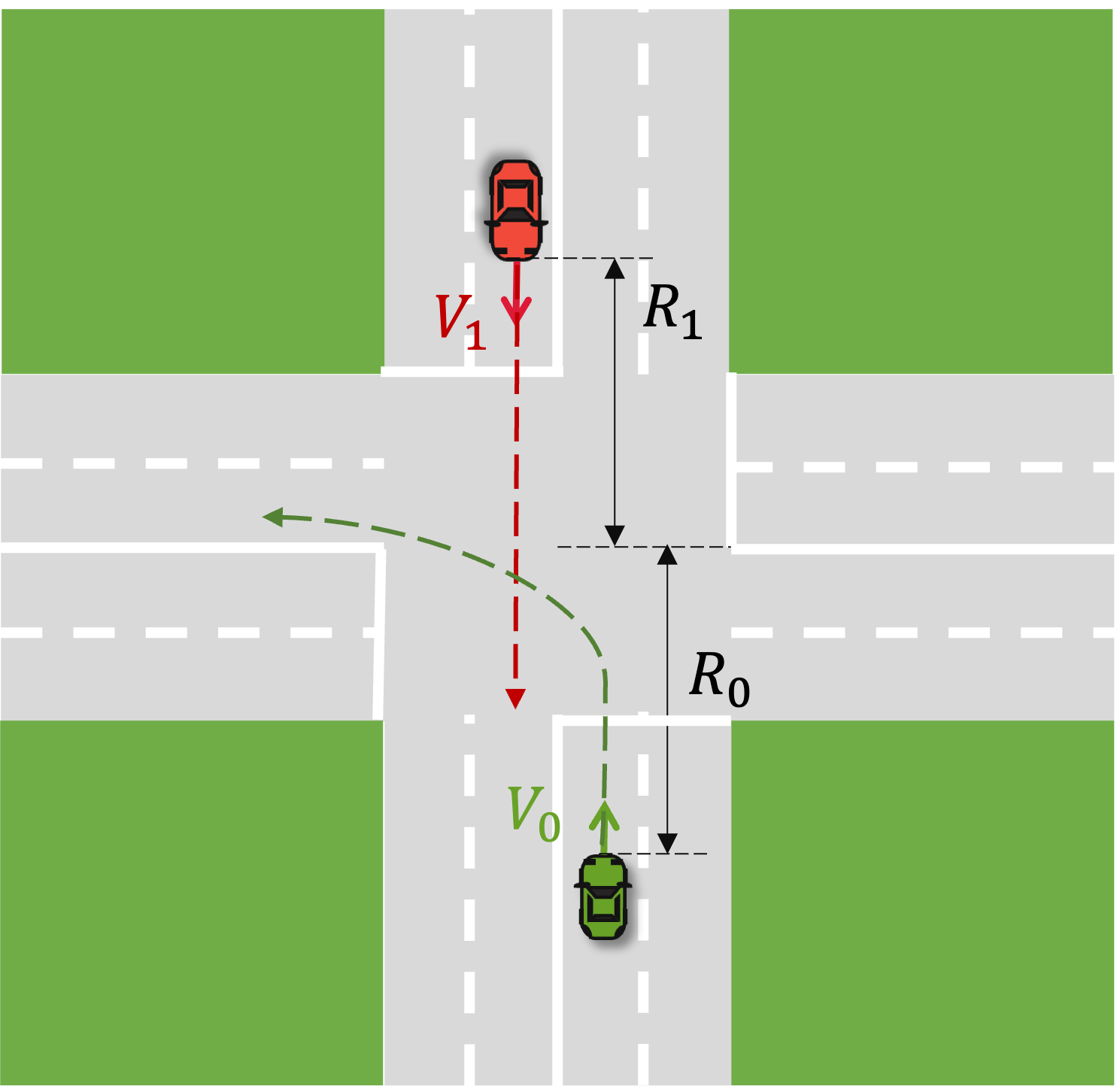}}
    \hspace{0.03\textwidth}
    \subfloat[NJRT]{\includegraphics[width=0.25\textwidth]{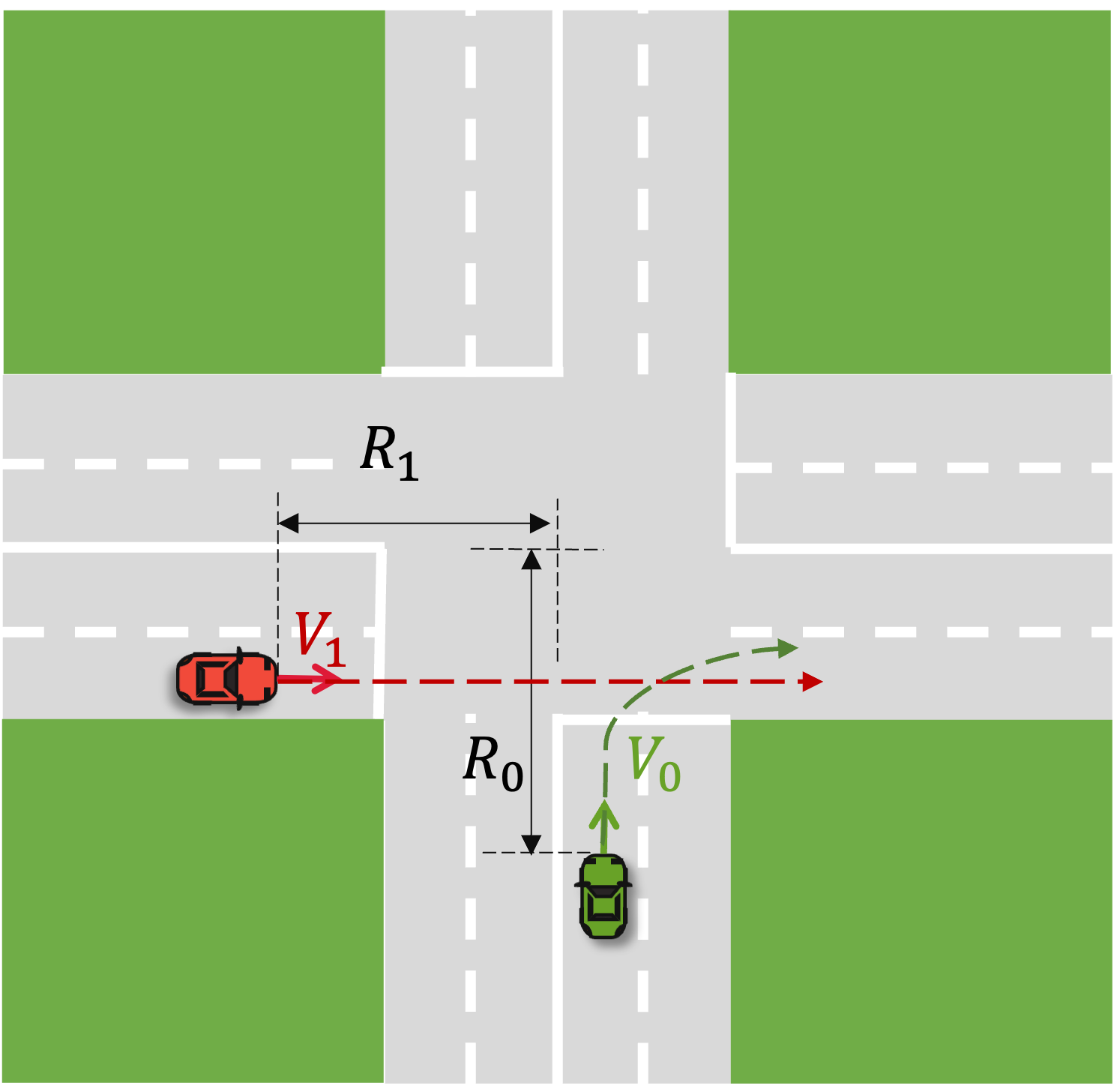}}
    \caption{Illustration of scenarios.}
    \label{scenarios-vis}
\end{figure}

The National Highway Traffic Safety Administration of the United States defines a series of pre-crash scenarios and calculates the average annual economic losses resulting from them ~\cite{najm2007pre}. For our experiment, we select the 6 scenarios that caused the highest economic losses each year and involve at least 2 vehicles.
They are illustrated in Figure ~\ref{scenarios-vis}:
\begin{itemize}[label={},leftmargin=*]
\item \textbf{FB: } The vehicle in front of the EGO suddenly brakes while driving.
\item \textbf{CutIn-1: }On the highway, the EGO is driving forward, and an NPC in the left lane cuts into the EGO's lane.
\item \textbf{CutIn-2: }Based on CutIn-1, another NPC is driving forward on the left rear of the EGO.
\item \textbf{OVTP: }Opposite vehicle taking priority. At an urban intersection without traffic lights, while the EGO goes straight from south to north, the NPC goes straight from west to east. 
\item \textbf{NJLT: }Non-signalized Junction Left Turn.  At an urban intersection without traffic lights, the EGO turns left and the NPC goes straight from north to south. 
\item \textbf{NJRT: }Non-signalized Junction Right Turn. At an urban intersection without traffic lights, the EGO turns right and the NPC goes straight from west to east. 
\end{itemize}
It is worth noting that the OSG framework can be applied to any logical scenario, not limited to the aforementioned 6 scenarios.

We compare OSG with the state-of-the-art BehAVExplor ~\cite{cheng2023behavexplor}, known as the latest and best approach for generating diverse and adversarial scenarios. Based on the idea of fuzz and utilizing energy-based mechanisms, BehAVExplor efficiently generates scenarios, surpassing the performance of representative algorithms such as AVFuzzer ~\cite{li2020av} and SAMOTA ~\cite{haq2022efficient}. It is noteworthy that the experiments of BehAVExplor were conducted on Baidu Apollo 6.0 and LGSVL 2021.3 simulators. However, the version of Baidu Apollo 6.0 is outdated, and testing this version is no longer practically relevant due to the frequent updates and iterations of the project. Moreover, the LGSVL simulator was sunset in 2022 and contains numerous bugs that will remain unresolved ~\cite{rong2020lgsvl}. In contrast, the Carla simulator is actively maintained. Consequently, we deploy  the BehAVExplor algorithm on Baidu Apollo 9.0 and Carla 0.9.14. 


\subsection{RQ1: Necessity of Variable-Risk Scenario Testing}

To answer this question, we will demonstrate whether scenarios with varying risk levels can effectively identify different types of potential risks. Additionally, we will explore whether integrating scenarios of different risk levels can provide an objective quantitative assessment of the safety of ADS.

\subsubsection{Collision Type Analysis}

\begin{figure}[t]
\centerline{\includegraphics[width=0.7\textwidth]{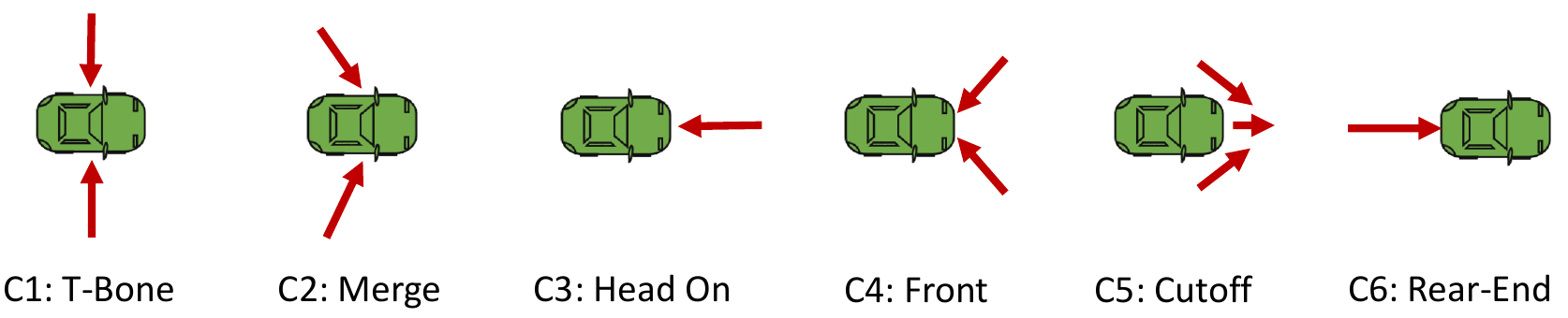}}
\caption{Collision types. The arrow indicates the position/direction of the NPC vehicle.}
\label{collision_types}
\end{figure}

\begin{table}[t]
  \caption{Collision types found by OSG under different $\omega$ settings. A larger value of $\omega$ indicates a higher risk level. The bold part denotes that this type did not appear in the test results across all risk levels. }
  \centering
  \label{collision-types-result}
\begin{tabular}{@{}ll@{}}
\toprule
$\omega$ & Collision Types                                                                                 \\ \midrule
0.0      & C1H,C1M,C1L,C2H,C2M,\textbf{C2L},\textbf{C3H},C4H,C4M,\textbf{C4L},C5H,C5M,\textbf{C6H},C6M,C6L \\
0.3      & C1H,C1M,C1L,C2H,C2M,\textbf{C3H},C4H,C4M,\textbf{C4L},C5H,C5M,\textbf{C5L},C6M,C6L              \\
0.5      & C1H,C1M,C1L,C2H,C2M,\textbf{C2L},\textbf{C3M},C4H,C4M,C5H,C5M,\textbf{C5L},\textbf{C6H},C6M,C6L \\
0.7      & C1H,C1M,C1L,C2H,C2M,\textbf{C2L},C4H,C4M,C5H,C5M,\textbf{C6H},C6M,C6L                           \\
1.0      & C1H,C1M,C1L,C2H,C2M,\textbf{C2L},\textbf{C3M},C4H,C4M,\textbf{C4L},C5H,C5M,\textbf{C5L},C6M,C6L \\ \bottomrule
\end{tabular}
\end{table}

Collisions can generally be classified into 6 types based on the relative positions of the vehicles at the time of impact ~\cite{rempe2022generating}, as illustrated in Figure ~\ref{collision_types} . We further categorize each main collision type into 3 sub-classes according to the relative velocity \(\Delta v\) between the NPC and the EGO. Subclass H indicates that \(\Delta v > 5 \text{m/s}\), subclass L indicates \(\Delta v < -5 \text{m/s}\), and subclass M indicates \(\Delta v \in [-5, 5] \, \text{m/s}\). For instance, C5L refers to a Cutoff collision where the NPC's velocity at the moment of collision is more than 5m/s lower than that of the EGO.

We integrate a data collector into the simulator to record the coordinates, speed, and heading angle of vehicles at the moment of collision. For Apollo and each of the 6 scenarios, we generate 5000 instances across 5 risk levels, repeating the process 10 times. Table~\ref{collision-types-result} enumerates the types of collisions occurring at each risk level.

Under different risk levels, the types of accidents in the generated scenarios vary. For example, when $\omega$=0.7, the collision type C6H is present, which is not found when $\omega$=1.
The C3M collision type only appears in scenarios with a medium or higher risk level, while C3H only appears in scenarios with a lower risk level. This finding suggests that scenarios with different risk levels exhibit distinct inherent vehicle behavior patterns.
This observation demonstrates that to thoroughly test ADS, it is necessary to comprehensively test scenarios across different risk levels to uncover a wider range of hidden issues.

\subsubsection{Quantitative Assessment of ADSs}

\begin{figure}[t]
\centerline{\includegraphics[width=0.7\textwidth]{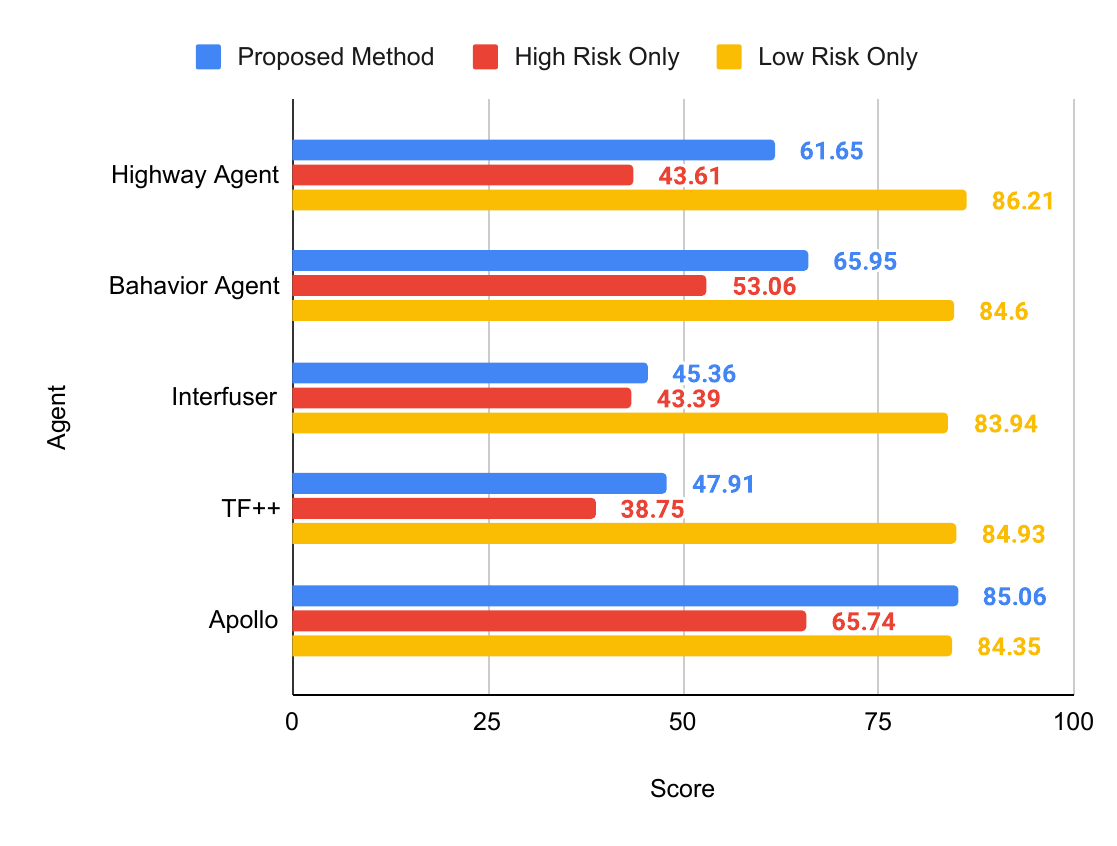}}
\caption{Evaluation results for different ADSs. \textbf{High Risk Only}: only $\omega=1.0$ scenarios are considered. \textbf{Low Risk Only}: only $\omega=0.0$ scenarios are considered.}
\label{score_ads}
\end{figure}

\begin{table}[t]
  \caption{Evaluation results for different versions of Apollo.}
  \centering
  \label{score_apollo}
\begin{tabular}{@{}llllllll@{}}
\toprule
\multirow{2}{*}{Apollo Version} & \multicolumn{6}{c}{Scenario Type}                                                                   & \multirow{2}{*}{Average} \\ \cmidrule(lr){2-7}
                                & FrontBrake     & CutIn-1        & CutIn-2        & OVTP           & NJLT           & NJRT           &                          \\ \midrule
9.0                             & 87.77          & 90.38          & 98.61          & \textbf{86.69} & \textbf{85.16} & 61.73          & \textbf{85.06}           \\
8.0                             & \textbf{89.12} & \textbf{91.31} & \textbf{98.74} & 85.64          & 76.73          & \textbf{61.77} & 83.89                    \\
7.0                             & 80.88          & 65.85          & 97.67          & 83.8           & 68.75          & 50.34          & 74.55                    \\ \bottomrule
\end{tabular}
\end{table}

The scenario test sets generated at different risk levels by OSG can objectively provide a comprehensive score for ADSs. To illustrate this point, we created a scenario test suite for each ADS in the Table ~\ref{agents-used}, containing various scenarios and risk levels, to compare the safety performance among different ADSs.

The algorithm of scoring is described as follows.
Label the logical scenarios in Figure ~\ref{scenarios-vis} sequentially as $S_1, S_2, ..., S_6$.
$\psi^{i,w}_j$ represents the $j$th concrete scenario in the search results for \(S_i\)  when $\omega=w$.
The number of scenarios per type is \( n_s = 20 \) . Let \( \Omega = \{0.0, 0.3, 0.5, 0.7, 1.0\} \) be the set of risk levels. We can then express the average score for scenario type $S_i$ under risk level $\omega$ as follows:
\begin{equation}
Q_{\omega}(S_i) = \frac{100}{n_s} \sum_{j=1}^{n_s} (1-\widetilde{ADV}(\psi^{i,w}_j)).
\end{equation}
The weight \( K(\omega) \) associated with each risk level \( \omega \) is given by the Gaussian function:
\begin{equation}
K(\omega) =  e^{-\frac{(x-\mu)^2}{2\sigma^2}}, \quad \mu = 0.5, \sigma = 0.1.
\end{equation}
The design of $K(\omega)$ indicates that scenarios that balance naturalness and criticalness are more valuable as references, whereas scenarios that are extremely natural or extremely adversarial have less reference value. The total score $C$ of an ADS is computed using a weighted average of the scores across all scenarios and risk levels. This can be written as:
\begin{equation}
Q = \frac{\sum_{\omega \in \Omega} K(\omega) \sum_{i=1}^{6} Q_{\omega}(S_i)}{ \sum_{i=1}^{6} \sum_{\omega \in \Omega} K(\omega)}.
\end{equation}
Here, the numerator represents the sum of weighted scenario scores across all types and risk levels, and the denominator is the sum of the weights for normalization. A higher score indicates better safety performance of ADS.

Figure ~\ref{score_ads} compares our proposed ADS scoring method with variants that consider only high-risk or low-risk scenarios. Our method ranks the ADSs in the order of: Apollo > Behavior Agent > Highway Agent > TF++ > Interfuser, which matches with practical observations. Apollo's robust planning and control (PnC) capabilities ensure high safety standards as an industrial-level ADS.

Rule-based ADSs outperform end-to-end agents like TF++ and Interfuser because they directly use real environment data from the simulator for PnC, while the latter rely on simulated sensors for perception. Imperfect perception modules in TF++ and Interfuser can generate incorrect information, lowering their overall safety performance. Our rating method objectively reflects this disparity.

When evaluating ADSs using only high-risk scenarios, all agents receive lower ratings, with TF++ replacing Interfuser in last place. Conversely, when scoring is based solely on low-risk scenarios, most ADSs improve their ratings, with Highway Agent surpassing Apollo. Highway Agent's robustness in safe, natural settings where long-term planning and prediction are not necessary, combined with its extensive domain knowledge, allows it to intelligently navigate and avoid obstacles.

However, concluding that Highway Agent has better safety than Apollo based on low-risk scenarios alone would be biased.
A comprehensive evaluation result showing the safety scores at different risk level would be perhaps more useful in practice. For instance, the users can then make an informed decision in choosing the ADS according to the risk level of the expected environment.

During the development of ADS, developers often need to quantitatively assess the differences between various ADS versions. We used OSG to score the safety of three different Apollo versions. 
Table ~\ref{score_apollo} demonstrates that Apollo's safety performance gradually improves across various scenarios as the version upgrades, with significant improvements from Apollo 7.0 to 8.0, especially in turning maneuvers. Apollo 8.0 fixed 8 control module issues and 52 planning module issues that exist in Apollo 7.0, specifically targeting steering control problems through the resolution of issues \#14842 and \#15061 ~\cite{issueApollo}. The minimal score improvements from Apollo 8.0 to 9.0 were anticipated, as Apollo 9.0 focused on extensive refactoring with minimal changes to the PnC algorithms ~\cite{release_apollo_9}.


\begin{mybox}
Answer to RQ1: Testing scenarios with varying risk levels is useful to uncover more potential issues and comprehensively evaluate the safety performance of ADS.
\end{mybox}

\subsection{RQ2: Can OSG Generate Scenarios at Different Risk Levels?}

\begin{figure}[t]
\centerline{\includegraphics[width=1\textwidth]{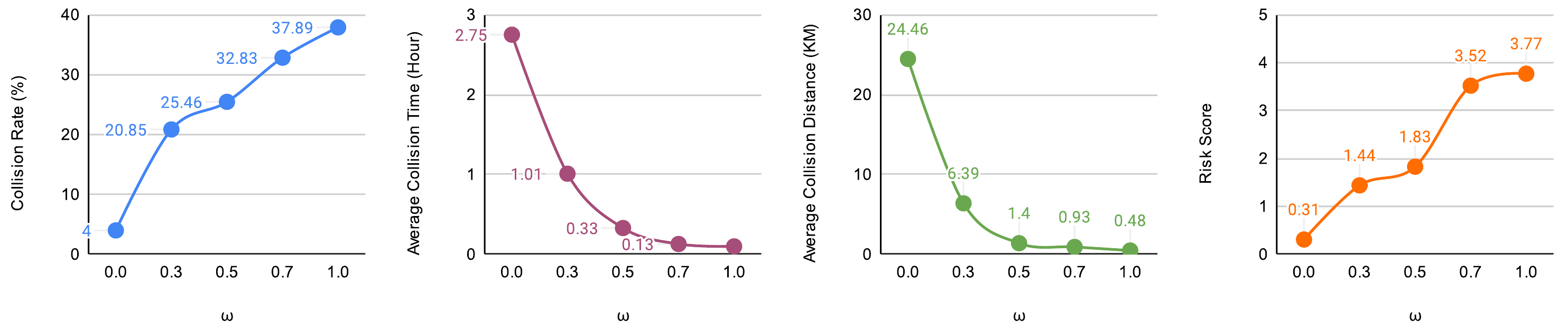}}
\caption{Risk Level Validation. The graphs from left to right illustrate the trends of Collision Rate, Average Collision Time, Average Collision Distance, and Ghodsi et al.'s score as $\omega$ increases.}
\label{trend_apollo}
\end{figure}

We generated 30,000 scenarios with Apollo using different logical scenarios and risk levels. Since there is no standard definition for a scenario's risk level, we adopted three widely used collision-related indicators ~\cite{niu2023re} : Collision Rate (CR) = number of collisions / number of scenarios, Average Collision Time (ACT) = total simulation time / number of collisions, and Average Collision Distance (ACD) = total EGO driven distance / number of collisions.

We also employed the algorithm proposed by Ghodsi et al. ~\cite{ghodsi2021generating} to evaluate the scenarios generated by OSG. Their method predicts the ADS's possible trajectories at every moment within a scenario, computes the risk probability based on each trajectory, and derives a risk score for the scenario. Despite its significant time complexity, this comprehensive assessment method is suitable for evaluating the risk level of generated scenarios.


Figure ~\ref{trend_apollo} demonstrates OSG's capability to generate traffic scenarios with varying risk levels by adjusting the $\omega$ parameter. As $\omega$ increases, there is a notable rise in CR and a decrease in both ACT and ACD, indicating more hazardous scenarios. The metrics by Ghodsi et al. further validate this trend, showing a direct correlation between $\omega$ and scenario risk intensity.





\begin{mybox}
Answer to RQ2: OSG can generate sets of scenarios according to the demand with different risk levels.
\end{mybox}

\subsection{RQ3: Effectiveness of OSG in Identifying Defects}

\begin{table*}[t]
  \caption{Comparison with BehAVExplor in terms of accident rate. OSG\_$C$ means the parameter of RIR $\omega=C$. BA: Behavior Agent. HA: Highway Agent.}
  \centering
  \label{all-compare}
\begin{tabular}{@{}llllllllll@{}}
\toprule
\multirow{2}{*}{Method}      & \multirow{2}{*}{Agents} & \multicolumn{6}{c}{Scenario Type}                                                               & \multirow{2}{*}{Avg.} & \multirow{2}{*}{\begin{tabular}[c]{@{}l@{}}Total\\ Avg.\end{tabular}} \\ \cmidrule(lr){3-8}
                             &                         & \begin{tabular}[c]{@{}l@{}}Front\\ Brake\end{tabular} & CutIn-1 & CutIn-2 & OVTP & NJLT & NJRT  &                       &                             \\ 

\midrule
\multirow{5}{*}{BehAVExplor} & Apollo                  & 1.4                                                   & 23.1    & 5.1     & 4.9  & 9.6  & 5     & 8.18                  & \multirow{5}{*}{14.10}      \\
                             & BA          & 40.1                                                  & 27.4    & 3.6     & 7.1  & 10.4 & 7.3   & 15.98                 &                             \\
                             & InterFuser              & 39.4                                                  & 31.3    & 8.4     & 7.9  & 9.7  & 7.5   & 17.37                 &                             \\
                             & TF++                    & 25.8                                                  & 27.5    & 15.9    & 7.2  & 12.3 & 5.3   & 15.67                 &                             \\
                             & HA           & 16.2                                                  & 33.4    & 7.7     & 6.3  & 9.5  & 6.7   & 13.30                 &                             \\ \cmidrule(lr){2-9}
\multirow{5}{*}{OSG\_1.0}    & Apollo                  & 50.4                                                  & 15.1    & 14.8    & 44.7 & 46.2 & 56.14 & 37.89                 & \multirow{5}{*}{\textbf{41.72}}      \\
                             & BA          & 49.8                                                  & 47.5    & 39.4    & 20.2 & 23.8 & 26.7  & 34.57                 &                             \\
                             & InterFuser              & 54.6                                                  & 67      & 53.8    & 58   & 40.8 & 62    & 56.03                 &                             \\
                             & TF++                    & 51                                                    & 46.9    & 48.5    & 49.5 & 48.7 & 57.7  & 50.38                 &                             \\
                             & HA           & 54.1                                                  & 38.6    & 29.8    & 14.2 & 21.7 & 20.1  & 29.75                 &                             \\ \cmidrule(lr){2-9}
\multirow{5}{*}{OSG\_0.7}    & Apollo                  & 47.95                                                 & 13.3    & 5.5     & 37.9 & 44.1 & 48.2  & 32.83                 & \multirow{5}{*}{34.05}      \\
                             & BA          & 42.7                                                  & 36      & 33      & 14.9 & 23.5 & 22.3  & 28.73                 &                             \\
                             & InterFuser              & 45.9                                                  & 52.8    & 42.3    & 41   & 40.6 & 51.8  & 45.73                 &                             \\
                             & TF++                    & 43.2                                                  & 26.1    & 30.3    & 47.4 & 36.5 & 50.7  & 39.03                 &                             \\
                             & HA           & 39.3                                                  & 32.9    & 25.8    & 11.4 & 19.4 & 14.8  & 23.93                 &                             \\ \cmidrule(lr){2-9}
\multirow{5}{*}{OSG\_0.5}    & Apollo                  & 34.6                                                  & 8.95    & 3.8     & 19   & 39.9 & 46.5  & 25.46                 & \multirow{5}{*}{29.28}      \\
                             & BA          & 25.8                                                  & 27      & 30.9    & 11.6 & 21.9 & 16.3  & 22.25                 &                             \\
                             & InterFuser              & 44.4                                                  & 52.6    & 40.5    & 36.9 & 40.5 & 46.7  & 43.60                 &                             \\
                             & TF++                    & 41.6                                                  & 25      & 24.9    & 39.4 & 33.6 & 40.7  & 34.20                 &                             \\
                             & HA           & 31.7                                                  & 27.4    & 23.3    & 11.3 & 18   & 13.7  & 20.90                 &                             \\ \cmidrule(lr){2-9}
\multirow{5}{*}{OSG\_0.3}    & Apollo                  & 28.2                                                  & 1.1     & 1       & 16.6 & 35.4 & 42.8  & 20.85                 & \multirow{5}{*}{23.96}      \\
                             & BA          & 23.3                                                  & 25.6    & 21.6    & 11.5 & 19.1 & 15    & 19.35                 &                             \\
                             & InterFuser              & 43.7                                                  & 44.3    & 17.8    & 32.9 & 30.3 & 32.8  & 33.63                 &                             \\
                             & TF++                    & 39.9                                                  & 11.5    & 13.2    & 36.4 & 31.2 & 38.4  & 28.43                 &                             \\
                             & HA           & 23.1                                                  & 26.5    & 21.1    & 11   & 16.4 & 7.1   & 17.53                 &                             \\ \cmidrule(lr){2-9}
\multirow{5}{*}{OSG\_0.0}    & Apollo                  & 10.7                                                  & 0.2     & 0.5     & 2.2  & 3.2  & 7.2   & 4.00                  & \multirow{5}{*}{7.03}       \\
                             & BA          & 11.6                                                  & 3.2     & 8.7     & 3.3  & 5.2  & 4.8   & 6.13                  &                             \\
                             & InterFuser              & 40                                                    & 6.8     & 12.4    & 2.6  & 3.4  & 4.8   & 11.67                 &                             \\
                             & TF++                    & 18.76                                                 & 2.2     & 5.7     & 2.9  & 6.4  & 5     & 6.83                  &                             \\
                             & HA           & 2.9                                                   & 4.3     & 5       & 6.1  & 11.1 & 9.7   & 6.52                  &                             \\ 
\bottomrule
\end{tabular}
\end{table*}

We generated 1,000 concrete scenarios for each ADS under each logical scenario using OSG and BehAVExplor. We repeated this process 10 times and averaged the results.
Table ~\ref{all-compare} presents the accident rates of the generated scenarios. A higher accident rate indicates greater effectiveness in identifying potential bugs within the ADS.
The results clearly shows that OSG, when set to $\omega>0.5$, consistently generates higher collision rates compared to BehAVExplor across all agents and scenarios.
The number of accident scenarios generated by BehAVExplor falls between OSG\_0.0 and OSG\_0.3, constituting 58.84\% of OSG\_0.3 and 33.79\% of OSG\_1.0. 
The accident rate for all scenarios generated by OSG is 27.21\%, which is a 92.97\% improvement over BehAVExplor.

We compared the efficiency of different methods in generating scenarios, focusing on the time consumed, which can be divided into simulation time and analysis time. Simulation time refers to the duration spent running scenarios within the simulator, while analysis time involves algorithm initialization, data processing, and logging. 
Simulation time primarily depends on the ADS and simulator and is independent of the scenario generation method. 
Under various settings of $\omega$, OSG consistently requires around 1.2 seconds per scenario for analysis time. In contrast, BehAVExplor's analysis time averages 5.2 seconds per scenario. This discrepancy arises because OSG relies on pre-trained models, such as the Naturalness Estimator, which complete their inference processes instantaneously. Conversely, BehAVExplor requires sequential processing of multiple algorithmic modules, resulting in slower speeds.


\begin{mybox}
Answer to RQ3: Under different $\omega$ settings, OSG effectively finds bugs in all kinds of ADS. The overall efficiency of accident scenario generation by OSG is approximately twice that of BehAVExplor. 
\end{mybox}

\subsection{Threats to Validity}

The OSG framework faces several threats, primarily from the simulation environment. 
First, conducting Apollo simulations within the Carla simulator poses potential threats due to the lack of official support from Apollo for Carla. The joint simulation bridge for Carla and Apollo, maintained by the Carla open-source community ~\cite{carla_apollo_bridge}, has issues such as data errors and high latency in the simulation of perception sensors. To circumvent this issue, we disabled Apollo's perception sensors and used the real data from the Carla simulator as the perception results sent to Apollo, ensuring more reliable simulations. This means that our main focus was testing Apollo's planning and control modules. 


Second, the selection of ADS and scenarios could pose a threat. To address this, we selected 6 typical traffic scenarios and 3 types of ADS for combination experiments, avoiding coincidental conclusions due to specific scenarios and ADS. 

Third, running Apollo, Interfuser, and TF++ requires significant performance, and long duration of simulation can easily lead to delays or even crashes. This may interrupt the generation process and potentially lead to incorrect data being recorded in the experiment results. To avoid this issue, we employed several methods: (1) We used multiple high-performance GPU servers for the experiments; (2) We implemented a watchdog module in the framework that periodically restarts ADS and the simulator. Moreover, if any module is detected to have a timeout response, the watchdog clears the process and restarts from the previous checkpoint.

Fourth, the traffic prior model is trained on specific datasets, which may limit its generalization. However, training on targeted data is appropriate given that regions differ in regulations and environmental conditions, necessitating local model adaptation. For example, right-hand drive data is unnecessary when developing systems for left-hand drive vehicles.

Lastly, the randomness of the experiment is a threat. The experiment involves multiple modules, including search algorithms and ADS, which all have a certain level of randomness. We counter this threat by setting a unified random seed parameter and conducting the experiment 10 times with different random seeds.

\section{Related Works}

Scenario-based simulation testing has become the primary method for testing Autonomous Driving Systems (ADS) ~\cite{lou2022testing}. However, applying traditional software testing methods to scenario testing is challenging due to the high dimension of variables and numerous dynamic entities in traffic scenarios. Researchers have proposed various scenario generation methods, including 
data-driven generation ~\cite{scanlon2021waymo, abdelfattah2021towards, chen2021geosim,savkin2021unsupervised,Guo2022LiRTestAL, Tian2022GeneratingCT,Deng2022ScenariobasedTR, xin2024litsimconflictawarepolicylongterm}, knowledge-based generation ~\cite{shiroshita2020behaviorally, wang2021commonroad, Zhang2023BuildingCT,Cao2021InvisibleFB, Cao2022YouCS, Chang2024LLMScenarioLL, Woodlief2024S3CSS}, and adversarial generation ~\cite{zhu2021adversarial, ding2021multimodal, ghodsi2021generating, arief2021deep, koren2018adaptive, cai2020summit,li2023simulation,Hildebrandt2023PhysCovPT, cheng2023behavexplor,Huai2023DoppelgngerTG,Zhong2021DetectingMF,tian2022mosat,Luo2021TargetingRV,Kim2022DriveFuzzDA,Song2023DiscoveringAD, Wang2024DanceOT}. 

Data-driven methods generate new data based on existing scenario datasets. For example, ~\cite{Hildebrandt2023PhysCovPT} transfers real-world accident data to scenarios for simulators. LiRTest ~\cite{Guo2022LiRTestAL} employs ADS-specific metamorphic relations and transforms perception data to detect errors in 3D object detection models. ~\cite{Tian2022GeneratingCT} generates scenarios by mining influential behavior patterns from datasets. However, these methods often suffer from the long-tail effect of data.

Knowledge-based approaches rely on expert experience and domain knowledge. ~\cite{zhang2023testing} use real-world traffic regulations as scenario constraints, while the CARLA Autonomous Driving Leaderboard~\cite{carla_leaderboard} scenarios are manually designed by experts. Although practical, these methods lack efficiency. S3C ~\cite{Woodlief2024S3CSS} creates a variety of natural scenarios by manually defining scene graphs.

Adversarial generation methods offer high efficiency and scalability. DoppelTest~\cite{Huai2023DoppelgngerTG} and FusED~\cite{Zhong2021DetectingMF} employ genetic algorithms to generate critical scenarios. ~\cite{Kim2022DriveFuzzDA} utilize a quality-guided fuzzer to uncover bugs in ADSs and simulators. ~\cite{Wang2024DanceOT} introduced a surrogate model into the fuzzing framework to improve the efficiency of critical scenario generation.

However, these methods focus on either scenario naturalness or criticalness, without providing flexibility in controlling both aspects. Our approach combines the advantages of data-driven methods and adversarial generation methods, enabling the generation of scenarios with varying risk levels according to specific testing requirements.

\section{Conclusion}

This paper introduces the On-demand Scenario Generation (OSG) framework for generating traffic scenarios with varying risk levels to test ADSs. OSG employs a Risk Intensity Regulator and a speciation-based particle swarm optimization algorithm to efficiently generate diverse scenarios. Experiments on Carla simulator and highway-env demonstrate OSG's ability to generate scenarios across a spectrum of risk levels and identify defects in rule-based, end-to-end, and hybrid ADS architectures. The results highlight the importance of testing ADS in scenarios with different risk levels to uncover potential issues. Furthermore, OSG provides a comprehensive scoring method for quantitatively assessing ADS safety performance. The proposed framework offers an efficient and thorough approach to testing ADSs, contributing to the development of safer and more reliable autonomous vehicles.

Future research may focus on incorporating more dynamic factors, such as vehicle and pedestrian flow, into the scenarios.

\section{Data Availability}
Both the source code of OSG and all experimental data are publicly available at ~\cite{my_data}.

\begin{acks}
This work was supported by the National Natural Science Foundation of China (NSFC) under Grants 62232008, 62032010, 62232014 and 62272377, the NSFC Youth Program under Grant 62402367, and the Joint Funds of the NSFC under Grant U24A20238.
\end{acks}

\bibliographystyle{ACM-Reference-Format}
\bibliography{my}


\begin{thebibliography}{63}


\ifx \showCODEN    \undefined \def \showCODEN     #1{\unskip}     \fi
\ifx \showDOI      \undefined \def \showDOI       #1{#1}\fi
\ifx \showISBNx    \undefined \def \showISBNx     #1{\unskip}     \fi
\ifx \showISBNxiii \undefined \def \showISBNxiii  #1{\unskip}     \fi
\ifx \showISSN     \undefined \def \showISSN      #1{\unskip}     \fi
\ifx \showLCCN     \undefined \def \showLCCN      #1{\unskip}     \fi
\ifx \shownote     \undefined \def \shownote      #1{#1}          \fi
\ifx \showarticletitle \undefined \def \showarticletitle #1{#1}   \fi
\ifx \showURL      \undefined \def \showURL       {\relax}        \fi
\providecommand\bibfield[2]{#2}
\providecommand\bibinfo[2]{#2}
\providecommand\natexlab[1]{#1}
\providecommand\showeprint[2][]{arXiv:#2}

\bibitem[Abdelfattah et~al\mbox{.}(2021)]%
        {abdelfattah2021towards}
\bibfield{author}{\bibinfo{person}{Mazen Abdelfattah}, \bibinfo{person}{Kaiwen Yuan}, \bibinfo{person}{Z~Jane Wang}, {and} \bibinfo{person}{Rabab Ward}.} \bibinfo{year}{2021}\natexlab{}.
\newblock \showarticletitle{Towards universal physical attacks on cascaded camera-lidar 3d object detection models}. In \bibinfo{booktitle}{\emph{2021 IEEE International Conference on Image Processing (ICIP)}}. IEEE, \bibinfo{pages}{3592--3596}.
\newblock
\urldef\tempurl%
\url{https://doi.org/10.1109/ICIP42928.2021.9506016}
\showDOI{\tempurl}


\bibitem[Arief et~al\mbox{.}(2021)]%
        {arief2021deep}
\bibfield{author}{\bibinfo{person}{Mansur Arief}, \bibinfo{person}{Zhiyuan Huang}, \bibinfo{person}{Guru Koushik~Senthil Kumar}, \bibinfo{person}{Yuanlu Bai}, \bibinfo{person}{Shengyi He}, \bibinfo{person}{Wenhao Ding}, \bibinfo{person}{Henry Lam}, {and} \bibinfo{person}{Ding Zhao}.} \bibinfo{year}{2021}\natexlab{}.
\newblock \showarticletitle{Deep probabilistic accelerated evaluation: A robust certifiable rare-event simulation methodology for black-box safety-critical systems}. In \bibinfo{booktitle}{\emph{International Conference on Artificial Intelligence and Statistics}}. PMLR, \bibinfo{pages}{595--603}.
\newblock


\bibitem[Baidu(2024a)]%
        {apollo}
\bibfield{author}{\bibinfo{person}{Baidu}.} \bibinfo{year}{2024}\natexlab{a}.
\newblock \bibinfo{title}{Baidu Apollo team (2024), Apollo: Open Source Autonomous Driving}.
\newblock \bibinfo{howpublished}{\url{https://github.com/ApolloAuto/apollo}}.
\newblock
\newblock
\shownote{Accessed: 2024-03-25}.


\bibitem[Baidu(2024b)]%
        {issueApollo}
\bibfield{author}{\bibinfo{person}{Baidu}.} \bibinfo{year}{2024}\natexlab{b}.
\newblock \bibinfo{title}{{Issues of Apollo}}.
\newblock \bibinfo{howpublished}{\url{https://github.com/ApolloAuto/apollo/issues}}.
\newblock
\newblock
\shownote{Accessed: 2024-04-08}.


\bibitem[Baidu(2024c)]%
        {release_apollo_9}
\bibfield{author}{\bibinfo{person}{Baidu}.} \bibinfo{year}{2024}\natexlab{c}.
\newblock \bibinfo{title}{{Release Note of Apollo 9.0}}.
\newblock \bibinfo{howpublished}{\url{https://github.com/ApolloAuto/apollo/releases}}.
\newblock
\newblock
\shownote{Accessed: 2024-04-08}.


\bibitem[Bock et~al\mbox{.}(2020)]%
        {inDdataset}
\bibfield{author}{\bibinfo{person}{Julian Bock}, \bibinfo{person}{Robert Krajewski}, \bibinfo{person}{Tobias Moers}, \bibinfo{person}{Steffen Runde}, \bibinfo{person}{Lennart Vater}, {and} \bibinfo{person}{Lutz Eckstein}.} \bibinfo{year}{2020}\natexlab{}.
\newblock \showarticletitle{The inD Dataset: A Drone Dataset of Naturalistic Road User Trajectories at German Intersections}. In \bibinfo{booktitle}{\emph{2020 IEEE Intelligent Vehicles Symposium (IV)}}. \bibinfo{pages}{1929--1934}.
\newblock
\urldef\tempurl%
\url{https://doi.org/10.1109/IV47402.2020.9304839}
\showDOI{\tempurl}


\bibitem[Brits et~al\mbox{.}(2002)]%
        {brits2002niching}
\bibfield{author}{\bibinfo{person}{Riaan Brits}, \bibinfo{person}{Andries~P Engelbrecht}, {and} \bibinfo{person}{Frans van~den Bergh}.} \bibinfo{year}{2002}\natexlab{}.
\newblock \showarticletitle{A niching particle swarm optimizer}. In \bibinfo{booktitle}{\emph{Proceedings of the 4th Asia-Pacific conference on simulated evolution and learning}}, Vol.~\bibinfo{volume}{2}. \bibinfo{pages}{692--696}.
\newblock


\bibitem[Cai et~al\mbox{.}(2020)]%
        {cai2020summit}
\bibfield{author}{\bibinfo{person}{Panpan Cai}, \bibinfo{person}{Yiyuan Lee}, \bibinfo{person}{Yuanfu Luo}, {and} \bibinfo{person}{David Hsu}.} \bibinfo{year}{2020}\natexlab{}.
\newblock \showarticletitle{Summit: A simulator for urban driving in massive mixed traffic}. In \bibinfo{booktitle}{\emph{2020 IEEE International Conference on Robotics and Automation (ICRA)}}. IEEE, \bibinfo{pages}{4023--4029}.
\newblock


\bibitem[Cao et~al\mbox{.}(2022)]%
        {Cao2022YouCS}
\bibfield{author}{\bibinfo{person}{Yulong Cao}, \bibinfo{person}{S.~Hrushikesh Bhupathiraju}, \bibinfo{person}{Pirouz Naghavi}, \bibinfo{person}{Takeshi Sugawara}, \bibinfo{person}{Zhuoqing~Morley Mao}, {and} \bibinfo{person}{Sara Rampazzi}.} \bibinfo{year}{2022}\natexlab{}.
\newblock \showarticletitle{You Can't See Me: Physical Removal Attacks on LiDAR-based Autonomous Vehicles Driving Frameworks}.
\newblock \bibinfo{journal}{\emph{ArXiv}}  \bibinfo{volume}{abs/2210.09482} (\bibinfo{year}{2022}).
\newblock
\urldef\tempurl%
\url{https://api.semanticscholar.org/CorpusID:252967801}
\showURL{%
\tempurl}


\bibitem[Cao et~al\mbox{.}(2021)]%
        {Cao2021InvisibleFB}
\bibfield{author}{\bibinfo{person}{Yulong Cao}, \bibinfo{person}{Ningfei Wang}, \bibinfo{person}{Chaowei Xiao}, \bibinfo{person}{Dawei Yang}, \bibinfo{person}{Jin Fang}, \bibinfo{person}{Ruigang Yang}, \bibinfo{person}{Qi~Alfred Chen}, \bibinfo{person}{Mingyan Liu}, {and} \bibinfo{person}{Bo Li}.} \bibinfo{year}{2021}\natexlab{}.
\newblock \showarticletitle{Invisible for both Camera and LiDAR: Security of Multi-Sensor Fusion based Perception in Autonomous Driving Under Physical-World Attacks}.
\newblock \bibinfo{journal}{\emph{2021 IEEE Symposium on Security and Privacy (SP)}} (\bibinfo{year}{2021}), \bibinfo{pages}{176--194}.
\newblock
\urldef\tempurl%
\url{https://doi.org/10.1109/SP40001.2021.00076}
\showDOI{\tempurl}


\bibitem[CARLA(2024)]%
        {carla_leaderboard}
\bibfield{author}{\bibinfo{person}{CARLA}.} \bibinfo{year}{2024}\natexlab{}.
\newblock \bibinfo{title}{{CARLA Autonomous Driving Leaderboard}}.
\newblock \bibinfo{howpublished}{\url{https://leaderboard.carla.org/}}.
\newblock
\newblock
\shownote{Accessed: 2024-03-25}.


\bibitem[Chang et~al\mbox{.}(2024)]%
        {Chang2024LLMScenarioLL}
\bibfield{author}{\bibinfo{person}{Chen Chang}, \bibinfo{person}{Siqi Wang}, \bibinfo{person}{Jiawei Zhang}, \bibinfo{person}{Jingwei Ge}, {and} \bibinfo{person}{Li Li}.} \bibinfo{year}{2024}\natexlab{}.
\newblock \showarticletitle{LLMScenario: Large Language Model Driven Scenario Generation}.
\newblock \bibinfo{journal}{\emph{IEEE Transactions on Systems, Man, and Cybernetics: Systems}} (\bibinfo{year}{2024}).
\newblock
\urldef\tempurl%
\url{https://doi.org/10.1109/TSMC.2024.3392930}
\showDOI{\tempurl}


\bibitem[Chen et~al\mbox{.}(2021)]%
        {chen2021geosim}
\bibfield{author}{\bibinfo{person}{Yun Chen}, \bibinfo{person}{Frieda Rong}, \bibinfo{person}{Shivam Duggal}, \bibinfo{person}{Shenlong Wang}, \bibinfo{person}{Xinchen Yan}, \bibinfo{person}{Sivabalan Manivasagam}, \bibinfo{person}{Shangjie Xue}, \bibinfo{person}{Ersin Yumer}, {and} \bibinfo{person}{Raquel Urtasun}.} \bibinfo{year}{2021}\natexlab{}.
\newblock \showarticletitle{Geosim: Realistic video simulation via geometry-aware composition for self-driving}. In \bibinfo{booktitle}{\emph{Proceedings of the IEEE/CVF conference on computer vision and pattern recognition}}. \bibinfo{pages}{7230--7240}.
\newblock


\bibitem[Cheng et~al\mbox{.}(2023)]%
        {cheng2023behavexplor}
\bibfield{author}{\bibinfo{person}{Mingfei Cheng}, \bibinfo{person}{Yuan Zhou}, {and} \bibinfo{person}{Xiaofei Xie}.} \bibinfo{year}{2023}\natexlab{}.
\newblock \showarticletitle{Behavexplor: Behavior diversity guided testing for autonomous driving systems}. In \bibinfo{booktitle}{\emph{Proceedings of the 32nd ACM SIGSOFT International Symposium on Software Testing and Analysis}}. \bibinfo{pages}{488--500}.
\newblock
\urldef\tempurl%
\url{https://doi.org/10.1145/3597926.3598072}
\showDOI{\tempurl}


\bibitem[Deng et~al\mbox{.}(2022)]%
        {Deng2022ScenariobasedTR}
\bibfield{author}{\bibinfo{person}{Yao Deng}, \bibinfo{person}{James~Xi Zheng}, \bibinfo{person}{Mengshi Zhang}, \bibinfo{person}{Guannan Lou}, {and} \bibinfo{person}{Tianyi Zhang}.} \bibinfo{year}{2022}\natexlab{}.
\newblock \showarticletitle{Scenario-based test reduction and prioritization for multi-module autonomous driving systems}.
\newblock \bibinfo{journal}{\emph{Proceedings of the 30th ACM Joint European Software Engineering Conference and Symposium on the Foundations of Software Engineering}} (\bibinfo{year}{2022}).
\newblock
\urldef\tempurl%
\url{https://doi.org/10.1145/3540250.3549152}
\showDOI{\tempurl}


\bibitem[Ding et~al\mbox{.}(2021)]%
        {ding2021multimodal}
\bibfield{author}{\bibinfo{person}{Wenhao Ding}, \bibinfo{person}{Baiming Chen}, \bibinfo{person}{Bo Li}, \bibinfo{person}{Kim~Ji Eun}, {and} \bibinfo{person}{Ding Zhao}.} \bibinfo{year}{2021}\natexlab{}.
\newblock \showarticletitle{Multimodal safety-critical scenarios generation for decision-making algorithms evaluation}.
\newblock \bibinfo{journal}{\emph{IEEE Robotics and Automation Letters}} \bibinfo{volume}{6}, \bibinfo{number}{2} (\bibinfo{year}{2021}), \bibinfo{pages}{1551--1558}.
\newblock


\bibitem[Dosovitskiy et~al\mbox{.}(2017)]%
        {Dosovitskiy17}
\bibfield{author}{\bibinfo{person}{Alexey Dosovitskiy}, \bibinfo{person}{German Ros}, \bibinfo{person}{Felipe Codevilla}, \bibinfo{person}{Antonio Lopez}, {and} \bibinfo{person}{Vladlen Koltun}.} \bibinfo{year}{2017}\natexlab{}.
\newblock \showarticletitle{{CARLA}: {An} Open Urban Driving Simulator}. In \bibinfo{booktitle}{\emph{Proceedings of the 1st Annual Conference on Robot Learning}}. \bibinfo{pages}{1--16}.
\newblock


\bibitem[Fang et~al\mbox{.}(2020)]%
        {fang2020augmented}
\bibfield{author}{\bibinfo{person}{Jin Fang}, \bibinfo{person}{Dingfu Zhou}, \bibinfo{person}{Feilong Yan}, \bibinfo{person}{Tongtong Zhao}, \bibinfo{person}{Feihu Zhang}, \bibinfo{person}{Yu Ma}, \bibinfo{person}{Liang Wang}, {and} \bibinfo{person}{Ruigang Yang}.} \bibinfo{year}{2020}\natexlab{}.
\newblock \showarticletitle{Augmented LiDAR simulator for autonomous driving}.
\newblock \bibinfo{journal}{\emph{IEEE Robotics and Automation Letters}} \bibinfo{volume}{5}, \bibinfo{number}{2} (\bibinfo{year}{2020}), \bibinfo{pages}{1931--1938}.
\newblock
\urldef\tempurl%
\url{https://doi.org/10.1109/LRA.2020.2969927}
\showDOI{\tempurl}


\bibitem[Ghodsi et~al\mbox{.}(2021)]%
        {ghodsi2021generating}
\bibfield{author}{\bibinfo{person}{Zahra Ghodsi}, \bibinfo{person}{Siva Kumar~Sastry Hari}, \bibinfo{person}{Iuri Frosio}, \bibinfo{person}{Timothy Tsai}, \bibinfo{person}{Alejandro Troccoli}, \bibinfo{person}{Stephen~W Keckler}, \bibinfo{person}{Siddharth Garg}, {and} \bibinfo{person}{Anima Anandkumar}.} \bibinfo{year}{2021}\natexlab{}.
\newblock \showarticletitle{Generating and characterizing scenarios for safety testing of autonomous vehicles}. In \bibinfo{booktitle}{\emph{2021 IEEE Intelligent Vehicles Symposium (IV)}}. IEEE, \bibinfo{pages}{157--164}.
\newblock
\urldef\tempurl%
\url{https://doi.org/10.1109/IV48863.2021.9576023}
\showDOI{\tempurl}


\bibitem[Guardstrikelab(2024)]%
        {carla_apollo_bridge}
\bibfield{author}{\bibinfo{person}{Guardstrikelab}.} \bibinfo{year}{2024}\natexlab{}.
\newblock \bibinfo{title}{Carla Apollo Bridge}.
\newblock \bibinfo{howpublished}{\url{https://github.com/guardstrikelab/carla_apollo_bridge}}.
\newblock


\bibitem[Guo et~al\mbox{.}(2022)]%
        {Guo2022LiRTestAL}
\bibfield{author}{\bibinfo{person}{An Guo}, \bibinfo{person}{Yang Feng}, {and} \bibinfo{person}{Zhenyu Chen}.} \bibinfo{year}{2022}\natexlab{}.
\newblock \showarticletitle{LiRTest: augmenting LiDAR point clouds for automated testing of autonomous driving systems}.
\newblock \bibinfo{journal}{\emph{Proceedings of the 31st ACM SIGSOFT International Symposium on Software Testing and Analysis}} (\bibinfo{year}{2022}).
\newblock
\urldef\tempurl%
\url{https://doi.org/10.1145/3533767.3534397}
\showDOI{\tempurl}


\bibitem[Hao et~al\mbox{.}(2023)]%
        {tivhkk}
\bibfield{author}{\bibinfo{person}{Kunkun Hao}, \bibinfo{person}{Wen Cui}, \bibinfo{person}{Yonggang Luo}, \bibinfo{person}{Lecheng Xie}, \bibinfo{person}{Yuqiao Bai}, \bibinfo{person}{Jucheng Yang}, \bibinfo{person}{Songyang Yan}, \bibinfo{person}{Yuxi Pan}, {and} \bibinfo{person}{Zijiang Yang}.} \bibinfo{year}{2023}\natexlab{}.
\newblock \showarticletitle{Adversarial Safety-Critical Scenario Generation using Naturalistic Human Driving Priors}.
\newblock \bibinfo{journal}{\emph{IEEE Transactions on Intelligent Vehicles}} (\bibinfo{year}{2023}), \bibinfo{pages}{1--16}.
\newblock
\urldef\tempurl%
\url{https://doi.org/10.1109/TIV.2023.3335862}
\showDOI{\tempurl}


\bibitem[Haq et~al\mbox{.}(2022)]%
        {haq2022efficient}
\bibfield{author}{\bibinfo{person}{Fitash~Ul Haq}, \bibinfo{person}{Donghwan Shin}, {and} \bibinfo{person}{Lionel Briand}.} \bibinfo{year}{2022}\natexlab{}.
\newblock \showarticletitle{Efficient online testing for dnn-enabled systems using surrogate-assisted and many-objective optimization}. In \bibinfo{booktitle}{\emph{Proceedings of the 44th international conference on software engineering}}. \bibinfo{pages}{811--822}.
\newblock
\urldef\tempurl%
\url{https://doi.org/10.1145/3510003.3510188}
\showDOI{\tempurl}


\bibitem[Hildebrandt et~al\mbox{.}(2023)]%
        {Hildebrandt2023PhysCovPT}
\bibfield{author}{\bibinfo{person}{Carl Hildebrandt}, \bibinfo{person}{Meriel von Stein}, {and} \bibinfo{person}{Sebastian~G. Elbaum}.} \bibinfo{year}{2023}\natexlab{}.
\newblock \showarticletitle{PhysCov: Physical Test Coverage for Autonomous Vehicles}.
\newblock \bibinfo{journal}{\emph{Proceedings of the 32nd ACM SIGSOFT International Symposium on Software Testing and Analysis}} (\bibinfo{year}{2023}).
\newblock
\urldef\tempurl%
\url{https://doi.org/10.1145/3597926.3598069}
\showDOI{\tempurl}


\bibitem[Huai et~al\mbox{.}(2023)]%
        {Huai2023DoppelgngerTG}
\bibfield{author}{\bibinfo{person}{Yuqi Huai}, \bibinfo{person}{Yuntianyi Chen}, \bibinfo{person}{Sumaya Almanee}, \bibinfo{person}{Tu~K Ngo}, \bibinfo{person}{Xiang Liao}, \bibinfo{person}{Ziwen Wan}, \bibinfo{person}{Qi~Alfred Chen}, {and} \bibinfo{person}{Joshua Garcia}.} \bibinfo{year}{2023}\natexlab{}.
\newblock \showarticletitle{Doppelg{\"a}nger Test Generation for Revealing Bugs in Autonomous Driving Software}.
\newblock \bibinfo{journal}{\emph{2023 IEEE/ACM 45th International Conference on Software Engineering (ICSE)}} (\bibinfo{year}{2023}), \bibinfo{pages}{2591--2603}.
\newblock
\urldef\tempurl%
\url{https://doi.org/10.1109/ICSE48619.2023.00216}
\showDOI{\tempurl}


\bibitem[Jaeger et~al\mbox{.}(2023)]%
        {jaeger2023hidden}
\bibfield{author}{\bibinfo{person}{Bernhard Jaeger}, \bibinfo{person}{Kashyap Chitta}, {and} \bibinfo{person}{Andreas Geiger}.} \bibinfo{year}{2023}\natexlab{}.
\newblock \showarticletitle{Hidden biases of end-to-end driving models}. In \bibinfo{booktitle}{\emph{Proceedings of the IEEE/CVF International Conference on Computer Vision}}. \bibinfo{pages}{8240--8249}.
\newblock


\bibitem[Kim et~al\mbox{.}(2022)]%
        {Kim2022DriveFuzzDA}
\bibfield{author}{\bibinfo{person}{Seulbae Kim}, \bibinfo{person}{Major Liu}, \bibinfo{person}{Junghwan~John Rhee}, \bibinfo{person}{Yuseok Jeon}, \bibinfo{person}{Yonghwi Kwon}, {and} \bibinfo{person}{Chung~Hwan Kim}.} \bibinfo{year}{2022}\natexlab{}.
\newblock \showarticletitle{DriveFuzz: Discovering Autonomous Driving Bugs through Driving Quality-Guided Fuzzing}.
\newblock \bibinfo{journal}{\emph{Proceedings of the 2022 ACM SIGSAC Conference on Computer and Communications Security}} (\bibinfo{year}{2022}).
\newblock
\urldef\tempurl%
\url{https://api.semanticscholar.org/CorpusID:253265101}
\showURL{%
\tempurl}


\bibitem[Koren et~al\mbox{.}(2018)]%
        {koren2018adaptive}
\bibfield{author}{\bibinfo{person}{Mark Koren}, \bibinfo{person}{Saud Alsaif}, \bibinfo{person}{Ritchie Lee}, {and} \bibinfo{person}{Mykel~J Kochenderfer}.} \bibinfo{year}{2018}\natexlab{}.
\newblock \showarticletitle{Adaptive stress testing for autonomous vehicles}. In \bibinfo{booktitle}{\emph{2018 IEEE Intelligent Vehicles Symposium (IV)}}. IEEE, \bibinfo{pages}{1--7}.
\newblock
\urldef\tempurl%
\url{https://doi.org/10.1109/ICRA40945.2020.9197228}
\showDOI{\tempurl}


\bibitem[Leurent(2018)]%
        {highway-env}
\bibfield{author}{\bibinfo{person}{Edouard Leurent}.} \bibinfo{year}{2018}\natexlab{}.
\newblock \bibinfo{title}{An Environment for Autonomous Driving Decision-Making}.
\newblock \bibinfo{howpublished}{\url{https://github.com/eleurent/highway-env}}.
\newblock


\bibitem[Li et~al\mbox{.}(2023)]%
        {li2023simulation}
\bibfield{author}{\bibinfo{person}{Changwen Li}, \bibinfo{person}{Joseph Sifakis}, \bibinfo{person}{Qiang Wang}, \bibinfo{person}{Rongjie Yan}, {and} \bibinfo{person}{Jian Zhang}.} \bibinfo{year}{2023}\natexlab{}.
\newblock \showarticletitle{Simulation-Based Validation for Autonomous Driving Systems}. In \bibinfo{booktitle}{\emph{Proceedings of the 32nd ACM SIGSOFT International Symposium on Software Testing and Analysis}}. \bibinfo{pages}{842--853}.
\newblock
\urldef\tempurl%
\url{https://doi.org/10.1145/3597926.3598100}
\showDOI{\tempurl}


\bibitem[Li et~al\mbox{.}(2020)]%
        {li2020av}
\bibfield{author}{\bibinfo{person}{Guanpeng Li}, \bibinfo{person}{Yiran Li}, \bibinfo{person}{Saurabh Jha}, \bibinfo{person}{Timothy Tsai}, \bibinfo{person}{Michael Sullivan}, \bibinfo{person}{Siva Kumar~Sastry Hari}, \bibinfo{person}{Zbigniew Kalbarczyk}, {and} \bibinfo{person}{Ravishankar Iyer}.} \bibinfo{year}{2020}\natexlab{}.
\newblock \showarticletitle{Av-fuzzer: Finding safety violations in autonomous driving systems}. In \bibinfo{booktitle}{\emph{2020 IEEE 31st international symposium on software reliability engineering (ISSRE)}}. IEEE, \bibinfo{pages}{25--36}.
\newblock
\urldef\tempurl%
\url{https://doi.org/10.1109/ISSRE5003.2020.00012}
\showDOI{\tempurl}


\bibitem[Lou et~al\mbox{.}(2022)]%
        {lou2022testing}
\bibfield{author}{\bibinfo{person}{Guannan Lou}, \bibinfo{person}{Yao Deng}, \bibinfo{person}{Xi Zheng}, \bibinfo{person}{Mengshi Zhang}, {and} \bibinfo{person}{Tianyi Zhang}.} \bibinfo{year}{2022}\natexlab{}.
\newblock \showarticletitle{Testing of autonomous driving systems: where are we and where should we go?}. In \bibinfo{booktitle}{\emph{Proceedings of the 30th ACM Joint European Software Engineering Conference and Symposium on the Foundations of Software Engineering}}. \bibinfo{pages}{31--43}.
\newblock
\urldef\tempurl%
\url{https://doi.org/10.1145/3540250.3549111}
\showDOI{\tempurl}


\bibitem[Luo et~al\mbox{.}(2021)]%
        {Luo2021TargetingRV}
\bibfield{author}{\bibinfo{person}{Yixing Luo}, \bibinfo{person}{Xiaoyi Zhang}, \bibinfo{person}{Paolo Arcaini}, \bibinfo{person}{Zhi Jin}, \bibinfo{person}{Haiyan Zhao}, \bibinfo{person}{Fuyuki Ishikawa}, \bibinfo{person}{Rongxin Wu}, {and} \bibinfo{person}{Tao Xie}.} \bibinfo{year}{2021}\natexlab{}.
\newblock \showarticletitle{Targeting Requirements Violations of Autonomous Driving Systems by Dynamic Evolutionary Search}.
\newblock \bibinfo{journal}{\emph{2021 36th IEEE/ACM International Conference on Automated Software Engineering (ASE)}} (\bibinfo{year}{2021}), \bibinfo{pages}{279--291}.
\newblock
\urldef\tempurl%
\url{https://doi.org/10.1109/ASE51524.2021.9678883}
\showDOI{\tempurl}


\bibitem[Manivasagam et~al\mbox{.}(2020)]%
        {manivasagam2020lidarsim}
\bibfield{author}{\bibinfo{person}{Sivabalan Manivasagam}, \bibinfo{person}{Shenlong Wang}, \bibinfo{person}{Kelvin Wong}, \bibinfo{person}{Wenyuan Zeng}, \bibinfo{person}{Mikita Sazanovich}, \bibinfo{person}{Shuhan Tan}, \bibinfo{person}{Bin Yang}, \bibinfo{person}{Wei-Chiu Ma}, {and} \bibinfo{person}{Raquel Urtasun}.} \bibinfo{year}{2020}\natexlab{}.
\newblock \showarticletitle{Lidarsim: Realistic lidar simulation by leveraging the real world}. In \bibinfo{booktitle}{\emph{Proceedings of the IEEE/CVF Conference on Computer Vision and Pattern Recognition}}. \bibinfo{pages}{11167--11176}.
\newblock


\bibitem[Minderhoud and Bovy(2001)]%
        {minderhoud2001extended}
\bibfield{author}{\bibinfo{person}{Michiel~M Minderhoud} {and} \bibinfo{person}{Piet~HL Bovy}.} \bibinfo{year}{2001}\natexlab{}.
\newblock \showarticletitle{Extended time-to-collision measures for road traffic safety assessment}.
\newblock \bibinfo{journal}{\emph{Accident Analysis \& Prevention}} \bibinfo{volume}{33}, \bibinfo{number}{1} (\bibinfo{year}{2001}), \bibinfo{pages}{89--97}.
\newblock
\urldef\tempurl%
\url{https://doi.org/10.1016/S0001-4575(00)00019-1}
\showDOI{\tempurl}


\bibitem[Najm et~al\mbox{.}(2007)]%
        {najm2007pre}
\bibfield{author}{\bibinfo{person}{Wassim~G Najm}, \bibinfo{person}{John~D Smith}, \bibinfo{person}{Mikio Yanagisawa}, {et~al\mbox{.}}} \bibinfo{year}{2007}\natexlab{}.
\newblock \bibinfo{booktitle}{\emph{Pre-crash scenario typology for crash avoidance research}}.
\newblock \bibinfo{type}{{T}echnical {R}eport}. \bibinfo{institution}{United States. Department of Transportation. National Highway Traffic Safety~…}.
\newblock


\bibitem[Niu et~al\mbox{.}(2023)]%
        {niu2023re}
\bibfield{author}{\bibinfo{person}{Haoyi Niu}, \bibinfo{person}{Kun Ren}, \bibinfo{person}{Yizhou Xu}, \bibinfo{person}{Ziyuan Yang}, \bibinfo{person}{Yichen Lin}, \bibinfo{person}{Yi Zhang}, {and} \bibinfo{person}{Jianming Hu}.} \bibinfo{year}{2023}\natexlab{}.
\newblock \showarticletitle{(Re)2H2O: Autonomous Driving Scenario Generation via Reversely Regularized Hybrid Offline-and-Online Reinforcement Learning}. In \bibinfo{booktitle}{\emph{2023 IEEE Intelligent Vehicles Symposium (IV)}}. \bibinfo{pages}{1--8}.
\newblock
\urldef\tempurl%
\url{https://doi.org/10.1109/IV55152.2023.10186559}
\showDOI{\tempurl}


\bibitem[Papamakarios et~al\mbox{.}(2017)]%
        {papamakarios2017masked}
\bibfield{author}{\bibinfo{person}{George Papamakarios}, \bibinfo{person}{Theo Pavlakou}, {and} \bibinfo{person}{Iain Murray}.} \bibinfo{year}{2017}\natexlab{}.
\newblock \showarticletitle{Masked autoregressive flow for density estimation}.
\newblock \bibinfo{journal}{\emph{Advances in neural information processing systems}}  \bibinfo{volume}{30} (\bibinfo{year}{2017}).
\newblock
\urldef\tempurl%
\url{https://doi.org/10.5555/3294771.3294994}
\showDOI{\tempurl}


\bibitem[Rempe et~al\mbox{.}(2022)]%
        {rempe2022generating}
\bibfield{author}{\bibinfo{person}{Davis Rempe}, \bibinfo{person}{Jonah Philion}, \bibinfo{person}{Leonidas~J Guibas}, \bibinfo{person}{Sanja Fidler}, {and} \bibinfo{person}{Or Litany}.} \bibinfo{year}{2022}\natexlab{}.
\newblock \showarticletitle{Generating useful accident-prone driving scenarios via a learned traffic prior}. In \bibinfo{booktitle}{\emph{Proceedings of the IEEE/CVF Conference on Computer Vision and Pattern Recognition}}. \bibinfo{pages}{17305--17315}.
\newblock


\bibitem[Rong et~al\mbox{.}(2020)]%
        {rong2020lgsvl}
\bibfield{author}{\bibinfo{person}{Guodong Rong}, \bibinfo{person}{Byung~Hyun Shin}, \bibinfo{person}{Hadi Tabatabaee}, \bibinfo{person}{Qiang Lu}, \bibinfo{person}{Steve Lemke}, \bibinfo{person}{M{\=a}rti{\c{n}}{\v{s}} Mo{\v{z}}eiko}, \bibinfo{person}{Eric Boise}, \bibinfo{person}{Geehoon Uhm}, \bibinfo{person}{Mark Gerow}, \bibinfo{person}{Shalin Mehta}, {et~al\mbox{.}}} \bibinfo{year}{2020}\natexlab{}.
\newblock \showarticletitle{LGSVL Simulator: A High Fidelity Simulator for Autonomous Driving}.
\newblock \bibinfo{journal}{\emph{arXiv preprint arXiv:2005.03778}} (\bibinfo{year}{2020}).
\newblock


\bibitem[Savkin et~al\mbox{.}(2021)]%
        {savkin2021unsupervised}
\bibfield{author}{\bibinfo{person}{Artem Savkin}, \bibinfo{person}{Rachid Ellouze}, \bibinfo{person}{Nassir Navab}, {and} \bibinfo{person}{Federico Tombari}.} \bibinfo{year}{2021}\natexlab{}.
\newblock \showarticletitle{Unsupervised traffic scene generation with synthetic 3D scene graphs}. In \bibinfo{booktitle}{\emph{2021 IEEE/RSJ International Conference on Intelligent Robots and Systems (IROS)}}. IEEE, \bibinfo{pages}{1229--1235}.
\newblock
\urldef\tempurl%
\url{https://doi.org/10.1109/IROS51168.2021.9636318}
\showDOI{\tempurl}


\bibitem[Scanlon et~al\mbox{.}(2021)]%
        {scanlon2021waymo}
\bibfield{author}{\bibinfo{person}{John~M Scanlon}, \bibinfo{person}{Kristofer~D Kusano}, \bibinfo{person}{Tom Daniel}, \bibinfo{person}{Christopher Alderson}, \bibinfo{person}{Alexander Ogle}, {and} \bibinfo{person}{Trent Victor}.} \bibinfo{year}{2021}\natexlab{}.
\newblock \showarticletitle{Waymo simulated driving behavior in reconstructed fatal crashes within an autonomous vehicle operating domain}.
\newblock \bibinfo{journal}{\emph{Accident Analysis \& Prevention}}  \bibinfo{volume}{163} (\bibinfo{year}{2021}), \bibinfo{pages}{106454}.
\newblock
\urldef\tempurl%
\url{https://doi.org/10.1016/j.aap.2021.106454}
\showDOI{\tempurl}


\bibitem[Shao et~al\mbox{.}(2023)]%
        {shao2023safety}
\bibfield{author}{\bibinfo{person}{Hao Shao}, \bibinfo{person}{Letian Wang}, \bibinfo{person}{Ruobing Chen}, \bibinfo{person}{Hongsheng Li}, {and} \bibinfo{person}{Yu Liu}.} \bibinfo{year}{2023}\natexlab{}.
\newblock \showarticletitle{Safety-enhanced autonomous driving using interpretable sensor fusion transformer}. In \bibinfo{booktitle}{\emph{Conference on Robot Learning}}. PMLR, \bibinfo{pages}{726--737}.
\newblock


\bibitem[Shiroshita et~al\mbox{.}(2020)]%
        {shiroshita2020behaviorally}
\bibfield{author}{\bibinfo{person}{Shinya Shiroshita}, \bibinfo{person}{Shirou Maruyama}, \bibinfo{person}{Daisuke Nishiyama}, \bibinfo{person}{Mario~Ynocente Castro}, \bibinfo{person}{Karim Hamzaoui}, \bibinfo{person}{Guy Rosman}, \bibinfo{person}{Jonathan DeCastro}, \bibinfo{person}{Kuan-Hui Lee}, {and} \bibinfo{person}{Adrien Gaidon}.} \bibinfo{year}{2020}\natexlab{}.
\newblock \showarticletitle{Behaviorally diverse traffic simulation via reinforcement learning}. In \bibinfo{booktitle}{\emph{2020 IEEE/RSJ International Conference on Intelligent Robots and Systems (IROS)}}. IEEE, \bibinfo{pages}{2103--2110}.
\newblock
\urldef\tempurl%
\url{https://doi.org/10.1109/IROS45743.2020.9341493}
\showDOI{\tempurl}


\bibitem[Siddiqui and Merrill(2023)]%
        {Siddiqui_Merrill_2023}
\bibfield{author}{\bibinfo{person}{Faiz Siddiqui} {and} \bibinfo{person}{Jeremy~B. Merrill}.} \bibinfo{year}{2023}\natexlab{}.
\newblock
\newblock
\urldef\tempurl%
\url{https://www.washingtonpost.com/technology/2023/06/10/tesla-autopilot-crashes-elon-musk/}
\showURL{%
\tempurl}


\bibitem[Song et~al\mbox{.}(2023)]%
        {Song2023DiscoveringAD}
\bibfield{author}{\bibinfo{person}{Ruoyu Song}, \bibinfo{person}{Muslum~Ozgur Ozmen}, \bibinfo{person}{Hyungsub Kim}, \bibinfo{person}{Raymond Muller}, \bibinfo{person}{Z.~Berkay Celik}, {and} \bibinfo{person}{Antonio Bianchi}.} \bibinfo{year}{2023}\natexlab{}.
\newblock \showarticletitle{Discovering Adversarial Driving Maneuvers against Autonomous Vehicles}. In \bibinfo{booktitle}{\emph{USENIX Security Symposium}}.
\newblock
\urldef\tempurl%
\url{https://api.semanticscholar.org/CorpusID:260777685}
\showURL{%
\tempurl}


\bibitem[Sun et~al\mbox{.}(2022)]%
        {sun2022lawbreaker}
\bibfield{author}{\bibinfo{person}{Yang Sun}, \bibinfo{person}{Christopher~M Poskitt}, \bibinfo{person}{Jun Sun}, \bibinfo{person}{Yuqi Chen}, {and} \bibinfo{person}{Zijiang Yang}.} \bibinfo{year}{2022}\natexlab{}.
\newblock \showarticletitle{LawBreaker: An Approach for Specifying Traffic Laws and Fuzzing Autonomous Vehicles}. In \bibinfo{booktitle}{\emph{37th IEEE/ACM International Conference on Automated Software Engineering}}. \bibinfo{pages}{1--12}.
\newblock


\bibitem[Thadani et~al\mbox{.}(2023)]%
        {Thadani_Lerman_Piper_Siddiqui_Uraizee_2023}
\bibfield{author}{\bibinfo{person}{Trisha Thadani}, \bibinfo{person}{Rachel Lerman}, \bibinfo{person}{Imogen Piper}, \bibinfo{person}{Faiz Siddiqui}, {and} \bibinfo{person}{Irfan Uraizee}.} \bibinfo{year}{2023}\natexlab{}.
\newblock
\newblock
\urldef\tempurl%
\url{https://www.washingtonpost.com/technology/interactive/2023/tesla-autopilot-crash-analysis/}
\showURL{%
\tempurl}


\bibitem[Tian et~al\mbox{.}(2022a)]%
        {tian2022mosat}
\bibfield{author}{\bibinfo{person}{Haoxiang Tian}, \bibinfo{person}{Yan Jiang}, \bibinfo{person}{Guoquan Wu}, \bibinfo{person}{Jiren Yan}, \bibinfo{person}{Jun Wei}, \bibinfo{person}{Wei Chen}, \bibinfo{person}{Shuo Li}, {and} \bibinfo{person}{Dan Ye}.} \bibinfo{year}{2022}\natexlab{a}.
\newblock \showarticletitle{MOSAT: finding safety violations of autonomous driving systems using multi-objective genetic algorithm}. In \bibinfo{booktitle}{\emph{Proceedings of the 30th ACM Joint European Software Engineering Conference and Symposium on the Foundations of Software Engineering}}. \bibinfo{pages}{94--106}.
\newblock
\urldef\tempurl%
\url{https://doi.org/10.1145/3540250.3549100}
\showDOI{\tempurl}


\bibitem[Tian et~al\mbox{.}(2022b)]%
        {Tian2022GeneratingCT}
\bibfield{author}{\bibinfo{person}{Haoxiang Tian}, \bibinfo{person}{Guoquan Wu}, \bibinfo{person}{Jiren Yan}, \bibinfo{person}{Yan Jiang}, \bibinfo{person}{Jun Wei}, \bibinfo{person}{W. Chen}, \bibinfo{person}{Shuo Li}, {and} \bibinfo{person}{Dan Ye}.} \bibinfo{year}{2022}\natexlab{b}.
\newblock \showarticletitle{Generating Critical Test Scenarios for Autonomous Driving Systems via Influential Behavior Patterns}.
\newblock \bibinfo{journal}{\emph{Proceedings of the 37th IEEE/ACM International Conference on Automated Software Engineering}} (\bibinfo{year}{2022}).
\newblock
\urldef\tempurl%
\url{https://api.semanticscholar.org/CorpusID:255441327}
\showURL{%
\tempurl}


\bibitem[Towers et~al\mbox{.}(2023)]%
        {towers_gymnasium_2023}
\bibfield{author}{\bibinfo{person}{Mark Towers}, \bibinfo{person}{Jordan~K. Terry}, \bibinfo{person}{Ariel Kwiatkowski}, \bibinfo{person}{John~U. Balis}, \bibinfo{person}{Gianluca~de Cola}, \bibinfo{person}{Tristan Deleu}, \bibinfo{person}{Manuel Goulão}, \bibinfo{person}{Andreas Kallinteris}, \bibinfo{person}{Arjun KG}, \bibinfo{person}{Markus Krimmel}, \bibinfo{person}{Rodrigo Perez-Vicente}, \bibinfo{person}{Andrea Pierré}, \bibinfo{person}{Sander Schulhoff}, \bibinfo{person}{Jun~Jet Tai}, \bibinfo{person}{Andrew Tan~Jin Shen}, {and} \bibinfo{person}{Omar~G. Younis}.} \bibinfo{year}{2023}\natexlab{}.
\newblock \bibinfo{title}{Gymnasium}.
\newblock
\newblock
\urldef\tempurl%
\url{https://doi.org/10.5281/zenodo.8127026}
\showDOI{\tempurl}


\bibitem[{U.S. Department of Transportation Federal Highway Administration}(2016)]%
        {usdot2016ngsim}
\bibfield{author}{\bibinfo{person}{{U.S. Department of Transportation Federal Highway Administration}}.} \bibinfo{year}{2016}\natexlab{}.
\newblock \bibinfo{title}{{Next Generation Simulation (NGSIM) Vehicle Trajectories and Supporting Data}}.
\newblock \bibinfo{howpublished}{[Dataset]. Provided by ITS DataHub through Data.transportation.gov}.
\newblock
\newblock
\shownote{Accessed 2023-05-13 from \url{http://doi.org/10.21949/1504477}}.


\bibitem[Wang et~al\mbox{.}(2021b)]%
        {wang2021advsim}
\bibfield{author}{\bibinfo{person}{Jingkang Wang}, \bibinfo{person}{Ava Pun}, \bibinfo{person}{James Tu}, \bibinfo{person}{Sivabalan Manivasagam}, \bibinfo{person}{Abbas Sadat}, \bibinfo{person}{Sergio Casas}, \bibinfo{person}{Mengye Ren}, {and} \bibinfo{person}{Raquel Urtasun}.} \bibinfo{year}{2021}\natexlab{b}.
\newblock \showarticletitle{Advsim: Generating safety-critical scenarios for self-driving vehicles}. In \bibinfo{booktitle}{\emph{Proceedings of the IEEE/CVF Conference on Computer Vision and Pattern Recognition}}. \bibinfo{pages}{9909--9918}.
\newblock


\bibitem[Wang et~al\mbox{.}(2024)]%
        {Wang2024DanceOT}
\bibfield{author}{\bibinfo{person}{Tong Wang}, \bibinfo{person}{Taotao Gu}, \bibinfo{person}{Huan Deng}, \bibinfo{person}{Hu Li}, \bibinfo{person}{Xiaohui Kuang}, {and} \bibinfo{person}{Gang Zhao}.} \bibinfo{year}{2024}\natexlab{}.
\newblock \showarticletitle{Dance of the ADS: Orchestrating Failures through Historically-Informed Scenario Fuzzing}.
\newblock \bibinfo{journal}{\emph{ArXiv}}  \bibinfo{volume}{abs/2407.04359} (\bibinfo{year}{2024}).
\newblock
\urldef\tempurl%
\url{https://api.semanticscholar.org/CorpusID:271039375}
\showURL{%
\tempurl}


\bibitem[Wang et~al\mbox{.}(2021a)]%
        {wang2021commonroad}
\bibfield{author}{\bibinfo{person}{Xiao Wang}, \bibinfo{person}{Hanna Krasowski}, {and} \bibinfo{person}{Matthias Althoff}.} \bibinfo{year}{2021}\natexlab{a}.
\newblock \showarticletitle{CommonRoad-RL: A configurable reinforcement learning environment for motion planning of autonomous vehicles}. In \bibinfo{booktitle}{\emph{2021 IEEE International Intelligent Transportation Systems Conference (ITSC)}}. IEEE, \bibinfo{pages}{466--472}.
\newblock
\urldef\tempurl%
\url{https://doi.org/10.1109/ITSC48978.2021.9564898}
\showDOI{\tempurl}


\bibitem[Woodlief et~al\mbox{.}(2024)]%
        {Woodlief2024S3CSS}
\bibfield{author}{\bibinfo{person}{Trey Woodlief}, \bibinfo{person}{Felipe Toledo}, \bibinfo{person}{Sebastian~G. Elbaum}, {and} \bibinfo{person}{Matthew~B. Dwyer}.} \bibinfo{year}{2024}\natexlab{}.
\newblock \showarticletitle{S3C: Spatial Semantic Scene Coverage for Autonomous Vehicles}.
\newblock \bibinfo{journal}{\emph{2024 IEEE/ACM 46th International Conference on Software Engineering (ICSE)}} (\bibinfo{year}{2024}), \bibinfo{pages}{1736--1748}.
\newblock
\urldef\tempurl%
\url{https://api.semanticscholar.org/CorpusID:269133131}
\showURL{%
\tempurl}


\bibitem[Xin et~al\mbox{.}(2024)]%
        {xin2024litsimconflictawarepolicylongterm}
\bibfield{author}{\bibinfo{person}{Haojie Xin}, \bibinfo{person}{Xiaodong Zhang}, \bibinfo{person}{Renzhi Tang}, \bibinfo{person}{Songyang Yan}, \bibinfo{person}{Qianrui Zhao}, \bibinfo{person}{Chunze Yang}, \bibinfo{person}{Wen Cui}, {and} \bibinfo{person}{Zijiang Yang}.} \bibinfo{year}{2024}\natexlab{}.
\newblock \bibinfo{title}{LitSim: A Conflict-aware Policy for Long-term Interactive Traffic Simulation}.
\newblock
\newblock
\showeprint[arxiv]{2403.04299}~[cs.RO]
\urldef\tempurl%
\url{https://arxiv.org/abs/2403.04299}
\showURL{%
\tempurl}


\bibitem[Yan et~al\mbox{.}(2024)]%
        {my_data}
\bibfield{author}{\bibinfo{person}{Songyang Yan}, \bibinfo{person}{Xiaodong Zhang}, \bibinfo{person}{Kunkun Hao}, \bibinfo{person}{Haojie Xin}, \bibinfo{person}{Yonggang Luo}, \bibinfo{person}{Jucheng Yang}, \bibinfo{person}{Ming Fan}, \bibinfo{person}{Chao Yang}, \bibinfo{person}{Jun Sun}, \bibinfo{person}{}, {and} \bibinfo{person}{Zijiang Yang}.} \bibinfo{year}{2024}\natexlab{}.
\newblock \bibinfo{title}{{OSG DATA at FSE 25}}.
\newblock
\newblock
\urldef\tempurl%
\url{https://doi.org/10.5281/zenodo.13621782}
\showDOI{\tempurl}


\bibitem[Zhan et~al\mbox{.}(2019)]%
        {zhan2019interaction}
\bibfield{author}{\bibinfo{person}{Wei Zhan}, \bibinfo{person}{Liting Sun}, \bibinfo{person}{Di Wang}, \bibinfo{person}{Haojie Shi}, \bibinfo{person}{Aubrey Clausse}, \bibinfo{person}{Maximilian Naumann}, \bibinfo{person}{Julius Kummerle}, \bibinfo{person}{Hendrik Konigshof}, \bibinfo{person}{Christoph Stiller}, \bibinfo{person}{Arnaud de La~Fortelle}, {et~al\mbox{.}}} \bibinfo{year}{2019}\natexlab{}.
\newblock \showarticletitle{Interaction dataset: An international, adversarial and cooperative motion dataset in interactive driving scenarios with semantic maps}.
\newblock \bibinfo{journal}{\emph{arXiv preprint arXiv:1910.03088}} (\bibinfo{year}{2019}).
\newblock


\bibitem[Zhang and Cai(2023)]%
        {Zhang2023BuildingCT}
\bibfield{author}{\bibinfo{person}{Xudong Zhang} {and} \bibinfo{person}{Yan Cai}.} \bibinfo{year}{2023}\natexlab{}.
\newblock \showarticletitle{Building Critical Testing Scenarios for Autonomous Driving from Real Accidents}.
\newblock \bibinfo{journal}{\emph{Proceedings of the 32nd ACM SIGSOFT International Symposium on Software Testing and Analysis}} (\bibinfo{year}{2023}).
\newblock
\urldef\tempurl%
\url{https://doi.org/10.1145/3597926.3598070}
\showDOI{\tempurl}


\bibitem[Zhang et~al\mbox{.}(2023)]%
        {zhang2023testing}
\bibfield{author}{\bibinfo{person}{Xiaodong Zhang}, \bibinfo{person}{Wei Zhao}, \bibinfo{person}{Yang Sun}, \bibinfo{person}{Jun Sun}, \bibinfo{person}{Yulong Shen}, \bibinfo{person}{Xuewen Dong}, {and} \bibinfo{person}{Zijiang Yang}.} \bibinfo{year}{2023}\natexlab{}.
\newblock \showarticletitle{Testing automated driving systems by breaking many laws efficiently}. In \bibinfo{booktitle}{\emph{Proceedings of the 32nd ACM SIGSOFT International Symposium on Software Testing and Analysis}}. \bibinfo{pages}{942--953}.
\newblock
\urldef\tempurl%
\url{https://doi.org/10.1145/3597926.3598108}
\showDOI{\tempurl}


\bibitem[Zhong et~al\mbox{.}(2021)]%
        {Zhong2021DetectingMF}
\bibfield{author}{\bibinfo{person}{Ziyuan Zhong}, \bibinfo{person}{Zhisheng Hu}, \bibinfo{person}{Shengjian Guo}, \bibinfo{person}{Xinyang Zhang}, \bibinfo{person}{Zhenyu Zhong}, {and} \bibinfo{person}{Baishakhi Ray}.} \bibinfo{year}{2021}\natexlab{}.
\newblock \showarticletitle{Detecting multi-sensor fusion errors in advanced driver-assistance systems}.
\newblock \bibinfo{journal}{\emph{Proceedings of the 31st ACM SIGSOFT International Symposium on Software Testing and Analysis}} (\bibinfo{year}{2021}).
\newblock
\urldef\tempurl%
\url{https://api.semanticscholar.org/CorpusID:246431152}
\showURL{%
\tempurl}


\bibitem[Zhu et~al\mbox{.}(2021)]%
        {zhu2021adversarial}
\bibfield{author}{\bibinfo{person}{Yi Zhu}, \bibinfo{person}{Chenglin Miao}, \bibinfo{person}{Foad Hajiaghajani}, \bibinfo{person}{Mengdi Huai}, \bibinfo{person}{Lu Su}, {and} \bibinfo{person}{Chunming Qiao}.} \bibinfo{year}{2021}\natexlab{}.
\newblock \showarticletitle{Adversarial attacks against lidar semantic segmentation in autonomous driving}. In \bibinfo{booktitle}{\emph{Proceedings of the 19th ACM conference on embedded networked sensor systems}}. \bibinfo{pages}{329--342}.
\newblock
\urldef\tempurl%
\url{https://doi.org/10.1145/3485730.3485935}
\showDOI{\tempurl}


\end{thebibliography}

\end{document}